\documentclass[10pt,journal,twocolumn]{IEEEtran}
\ifCLASSOPTIONcompsoc
  \usepackage[nocompress]{cite}
\else
  \usepackage{cite}
\fi

\ifCLASSINFOpdf
\else
\fi

\usepackage{times}
\usepackage{epsfig}
\usepackage{graphicx}
\usepackage{amsmath}
\usepackage{amssymb}
\usepackage{eucal,bibspacing}
\usepackage{cite}
\usepackage{cs}
\usepackage{mathsymb}
\usepackage{textcomp}
\usepackage{verbatim}
\usepackage{url}
\usepackage[super]{nth}
\usepackage{colortbl}
\usepackage{color}
\usepackage{xcolor}
\usepackage{mathtools}

\usepackage{multirow}
\usepackage{pbox}
\usepackage{wrapfig}
\usepackage{caption}
\usepackage[linesnumbered,algoruled,boxed,lined]{algorithm2e}

\let\savedalgorithm\algorithm
\let\savedendalgorithm\endalgorithm

\usepackage{algorithm}

\newcommand{\ie}{\textit{i.e., }}

\def\bf{{\boldsymbol f}}

\begin{document}

\title{Cross-Entropy Adversarial View Adaptation for Person Re-identification}

\author{Lin Wu, Richang Hong, Yang Wang*, Meng Wang 
\IEEEcompsocitemizethanks{\IEEEcompsocthanksitem \protect\\ Lin Wu is with the School of Computer Science and Information Engineering, Hefei University of Technology, Hefei 230000, China.
\protect\\ Richang Hong is with the School of Computer Science and Information Engineering, Hefei University of Technology, Hefei 230000, China. Email: hongrc.hfut@gmail.com.
\protect\\ Yang Wang (Corresponding author) is with the School of Computer Science and Information Engineering, Hefei University of Technology, Hefei 230000, China; Dalian University of Technology, Dalian 116024, China. Email: yangwang@hfut.edu.cn. Yang Wang was supported by National Natural Science Foundation of China, under Grant No 61806035.
\protect\\ Meng Wang is with the School of Computer Science and Information Engineering, Hefei University of Technology, Hefei 230000, China. Email: eric.mengwang@gmail.com. Meng Wang was supported by he National Key Research and Development Program of China under grant 2018YFB0804200; NSFC 61725203, 61732008.
}
}

\IEEEtitleabstractindextext{%
\begin{abstract}
Person re-identification (re-ID) is a task of matching pedestrians under disjoint camera views. To recognise paired snapshots, it has to cope with large cross-view variations caused by the camera view shift. Supervised deep neural networks are effective in producing a set of non-linear projections that can transform cross-view images into a common feature space. However, they typically impose a symmetric architecture, yielding the network ill-conditioned on its optimisation. In this paper, we learn view-invariant subspace for person re-ID, and its corresponding similarity metric using an adversarial view adaptation approach. The main contribution is to learn coupled asymmetric mappings regarding view characteristics which are adversarially trained to address the view discrepancy by optimising the cross-entropy view confusion objective. To determine the similarity value, the network is empowered with a similarity discriminator to promote features that are highly discriminant in distinguishing positive and negative pairs. The other contribution includes an adaptive weighing on the most difficult samples to address the imbalance of within/between-identity pairs. Our approach achieves notable improved performance in comparison to state-of-the-arts on benchmark datasets.
\end{abstract}

\begin{IEEEkeywords}
Person re-identification, View adaptation, Adversarial learning, Entropy regularisation.
\end{IEEEkeywords}}

\maketitle

\IEEEdisplaynontitleabstractindextext

\IEEEpeerreviewmaketitle

\section{Introduction}\label{sec:intro}

\IEEEPARstart{P}{erson} re-identification (re-ID) is a challenging problem specialising on pedestrian matching across a network of cameras. It has not been solved yet principally because of the significant visual changes caused by colour, background, camera viewpoints and human poses. Recent state-of-the-arts are developed in the basis of supervised deep neural networks \cite{DG-Dropout,Deep-Embed,What-and-where,SpindleNet,Part-Aligned,MSCAN,PIE-reid,SI-CI,LDAFisherVector} to learn robust and discriminative representations against visual variations. However, training deep architectures requires a large number of labeled image pairs across multiple camera views, which is prohibitively expensive and not scalable to real-world scenarios. To combat that challenge, a number of semi/un-supervised methods have been developed \cite{One-shot-RE-ID,GenerativeSaliency,eSDC,t-LRDC,LSRO,CAMEL,OL-MANS,L1-graph}. Some of them attempt to seek feature invariance by designing robust hand-crafted features \cite{One-shot-RE-ID,eSDC,GenerativeSaliency}. However, without the supervision of labeled data, the discrimination and specificity apt to camera-pair changes are not captured. Also, unsupervised methods treat samples from different views indiscriminately, and the effect of view-specific inference is not considered. On the other hand, some unsupervised methods introduce graph structure or clustering centroid \cite{L1-graph,LSRO,CAMEL,OL-MANS} to keep visually similar people close in the projected space. Nonetheless, it is insufficient to explore the discriminative space as the learning of view-specific projections into a shared subspace is optimised independently.

Matching pedestrian snapshots across camera views (probe and gallery) can be achieved by seeking a common subspace therein, and jointly optimising a measure for each pair of cross-view images. Siamese networks with deep convolutions are demonstrated to hold promise in person re-ID \cite{What-and-where,FPNN,Wu-TMM} by learning a set of nonlinear transformations that align the correlation of layer activations in deep neural networks. However, Siamese networks have layer-wise equality constraints on deep layered representations, which are commonly imposed within convolutional networks through weight sharing. The idea of Siamese networks is to enforce the exact consistency between the probe and the gallery mapping, where the learning of symmetric transformations can reduce the number of parameters in the deep model. Unfortunately, this may induce the optimisation poorly conditioned because the same network must handle images from two disjoint distributions.

\begin{figure}[t]
\centering
\includegraphics[width=8.5cm,height=4.5cm]{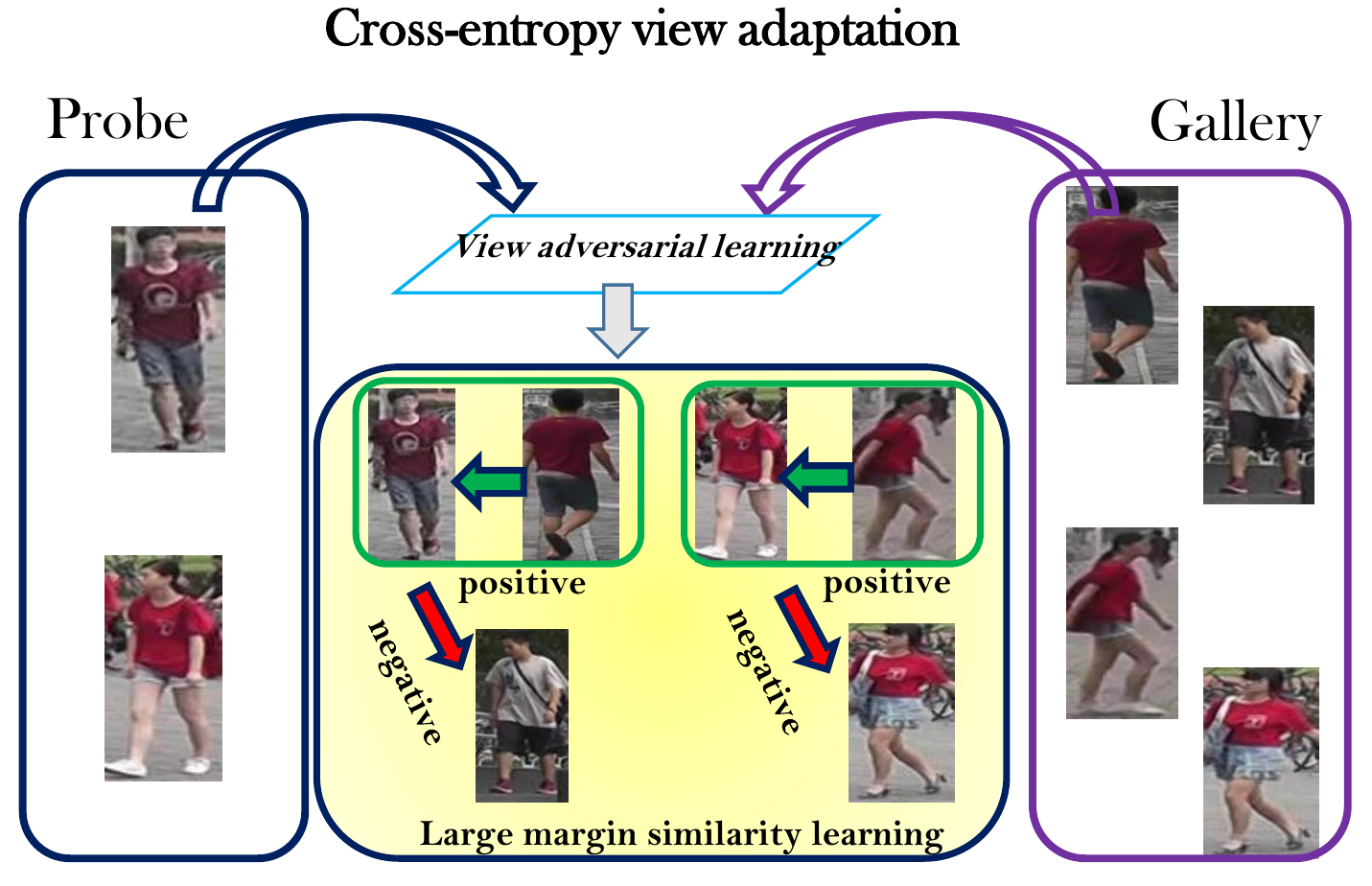}
\caption{Cross-entropy adversarial view adaptation for person re-ID. Given paired images across views (the probe and the gallery), the task is to jointly learn view-invariant feature space and its corresponding similarity metric. We propose to optimise asymmetric mappings regarding view specificity, and conditioned on each other through view-adversarial training (cross-entropy loss). The similarity metric is jointly learned by a discriminator network to identify positive pairs against negatives. See the text for details.}\label{fig:cross-adversarial}
\end{figure}

\subsection{Motivations}

This paper is motivated towards person re-ID by presenting a deep view adaptation approach in the sense that the non-linear transformations into a common feature space corresponding to paired observations should be asymmetric. This asymmetric architecture is necessary to characterise the view-specific entailments, and the optimisation regarding asymmetric mappings should be conditioned on each other to capture the identity interplay between the probe and gallery views. Also, a re-ID system proceeds paired images and requires a comparable metric to determine the similarity for each pair. Towards the above practices, in this paper we present a deep feature learning approach to optimise a feature space such that the invariance between the probe and gallery distribution is maximum (\textit{cross-view invariance}), and we jointly learn similarity metrics for paired images. Our approach is appealing in the ability to learn asymmetric mappings characterising cross-view images without enforcing any sharing constraints. The minimisation on view discrepancy is achieved by performing adversarial learning with entropy regularisation to operate cross-entropy minimisation over cross-view samples. To jointly learn a comparable metric, the adversarial framework is empowered with a discriminator network to distinguish positive pairs against negatives. More importantly, the network training does not require a large number of training samples as opposed to existing deep learning methods \cite{Part-Aligned,FPNN,Multi-loss} because we introduce adaptive weighting into the paired inputs which would emphasise the most difficult ones by assigning batch-based adaptive weights into positive/negative pairs.

It is noted that our framework is different from the study on domain adaptation to person re-ID \cite{SPGAN,DG-Dropout,Re-id-ICDSC,Transfer-re-id}. First, this line typically reuses pre-trained models from a closely related dataset with a large amount of samples (source), and then design the training towards the much smaller dataset of interest (target). Here instead, we are interested in adapting adversarial learning into cross-view invariant feature learning a.k.a \textit{adversarial view adaptation}, to effectively address the view discrepancy in re-identifying persons. Secondly, a common problem of existing domain adaptation approaches is that a principled alignment between the source and target is missing, and thus they are unable to penalise the correlated domain misalignment in practical terms. In contrast, our method explicitly minimises the view discrepancy through the proposed view-adversarial objective. Our method is also distinct from existing methods based on adversarial losses \cite{SPGAN}. For instance, SPGAN \cite{SPGAN} is composed of GAN loss to update the target domain w.r.t the source, our method instead uses cross-entropy loss to optimise the view confusion objective.

\subsection{Our Approach and Contributions}

Our approach is designed based on the view adaptation scheme to learn asymmetric deep neural transformations in order to map view-specific distributions into a common feature space. In this sense, we introduce adversarial learning \cite{GAN} into \textit{view discriminator} which is optimised through cross-entropy based \textit{view confusion objective}. This objective is to confuse the view discriminator that will perceive the two distributions identically so as to minimise the cross-view discrepancy. Specifically, we develop an adaptive learning framework to produce asymmetric mappings over two views through a view-adversarial training. When the view discriminator cannot determine if a pair is from the probe or the gallery view, the feature inference becomes optimal in terms of creating a view-invariant space. In this adversarial learning regime, view adaptation is seen as a generative adversarial network but there is not necessary to generate samples. In fact, the discriminator is confused by a view confusion objective and cannot determine the samples are from the probe or the gallery distribution when the network is optimal in creating a latent space conditioned on two views.

To address the similarity learning, we additionally enforce the semantic similarity by learning a distance metric jointly with the feature learning. This similarity is pertained in the discriminative base model of adversarial networks by using a contrastive loss (i.e., \textit{similarity discriminator}), which pulls the images in positive pairs closer while pushes the negative pairs away from positives. Thus, our network is end-to-end trainable to process paired samples and outputs its similarity value to determine whether the pair is from the same identity or not. The overview of our framework is shown in Fig. \ref{fig:cross-adversarial}. However, training paired input would raise an imbalance issue between the within-identity and between-identity samples. Hence, we particular introduce adaptive weighing into the most difficult positive/negative ones, which leads to optimised re-ID rank loss and quick convergence \cite{MTMCT}.

The major contributions of this paper can be summarised as follows.

\begin{itemize}
  \item We propose a principled adversarial feature learning approach to person re-ID to jointly produce a latent view-invariant feature space and its corresponding distance metric which maintains high discrimination on positive pairs from negatives.
  \item Our method is differentiated from the literature in conceptualising cross-view matching through \textbf{asymmetric mappings} followed by explicit view adaptation. This is achieved by presenting view-adversarial learning to train cross-view embedding whilst confusing a view discriminator in a cross-entropy objective function.
  \item We provide insights into adaptive weighing which assigns larger weights to difficult samples such that positive/negative class imbalance is effectively addressed.
  \item Extensive validations of the proposed method against the state-of-the-art are performed to demonstrate the competence of our model.
\end{itemize}

\section{Related Work}\label{sec:related}

\subsection{Person Re-identification}
Most of existing re-ID models are developed in a supervised manner to learn discriminative features \cite{Deep-Embed,PIE-reid,Part-Aligned,VideoPerson,3D-PersonVLAD} or learn distance metrics \cite{NullSpace-Reid,SI-CI,Multi-loss}. However, these models commonly rely on substantial labeled training data, which would hinder the application of them in large networked cameras. Semi-supervised and unsupervised methods are presented to overcome the scalability issue by using limited number of labeled samples or without using label information. These techniques often focus on designing handcrafted features (e.g., colour, texture) \cite{One-shot-RE-ID,GenerativeSaliency,eSDC} that should be robust to visual changes in imaging conditions. However, low-level features are not expressive in view-invariance because features are not learned to be apt to view-specific bias. On the other hand, transfer learning has been applied into re-ID \cite{UMDL,PUL,t-LRDC,CAMEL,One-shot-RE-ID}, and these methods learn the model using large labeled datasets and then transfer the discriminative knowledge to the unlabelled target pairs. For example, they can learn a cross-view metric either by asymmetric clustering on person images \cite{CAMEL} or by transferable colour metric from a single pair of images \cite{One-shot-RE-ID}. However, there is still a considerable performance gap relative to supervised learning approaches because it is not principled to fully explore the discriminative space in the context of source and domain image translation. In contrast to existing approaches that derive a metric independently from images of people, we learn a deep metric jointly with feature learning from few labeled training pairs in an adversarial manner.

\subsection{Generative Adversarial Networks}

The Generative Adversarial Networks (GANs) \cite{GAN} consist of a generator $G$ and a discriminator $D$ to compete the learning where the generator is learned to map samples from a latent distribution to confuse $D$ by producing samples close to real data, while the discriminator tries to distinguish between real and generated samples. The most popular variation of GAN is the Deep Convolutional GAN (DCGAN) introduced by Radford et al. \cite{DCGAN}. DCGAN improved the overall quality of generated images by adapting Convolutional Neural Network (CNN) into GAN architecture. Then, GANs have been extensively studied and widely used in several applications including realistic image generation \cite{Generate-chairs}, image-image translation \cite{Cycle-GAN}, domain adaptation \cite{Few-shot-DA}, and cross-modal retrieval \cite{CYC-DGH}.

Recently, GANs are adopted into person re-ID community by Zhong \textit{et al} \cite{LSRO} which is to introduce a semi-supervised pipeline that integrates GAN-generated samples into the CNN learning. Following up the work in \cite{LSRO}, Deng \textit{et al} \cite{SPGAN} present an unsupervised domain adaptation method (SPGAN) to preserve the similarity after translation and then train re-ID models with the translated images using supervised feature learning methods. In this work, we do not focus on generating samples for person re-ID as opposed to \cite{LSRO}. Instead, we formulate adversarial networks into the feature learning process to produce a view-invariant subspace and jointly learning a similarity metric. Our method is also different from SPGAN \cite{SPGAN} in the sense that we learn asymmetric transformations regarding view discrepancy rather than addressing a target domain into matching the source domain.

\subsection{Adversarial Adaptation Methods}

Deep convolutional neural networks trained on large-scale datasets can learn representations which are generically useful across different tasks and visual domains \cite{Decaf,Cross-GANs-ReID}. However, due to the domain shift/bias, generalising the well-trained recognition models to novel tasks typically require fine-tuning these networks. While it is difficult to obtain enough labelled data to properly fine-tune the large number of parameters in deep networks, and thus recent deep adaptation methods attempt to mitigate the difficulty by learning deep neural transformations that map both domains into a common feature space. This can be generally achieved by optimising the representations to align the source and target sets \cite{Deep-CORAL,CORAL,MECA}. For instance, several methods use the maximum mean discrepancy loss to measure the difference between the source and the target feature distributions \cite{Deep-CORAL,Domain-confusion,DAN}. Inspired by the idea of adapting higher order statistics of the two distributions \cite{Link-ICDE-2013,Community-2013,Online-Multi-label,Outlier-links-2012}, some methods propose a transformation to minimise the distance between the covariance representations of source and target datasets to ultimately achieve the correlation alignment \cite{CORAL,MECA}. These approaches are unsupervised domain adaptation that do not need any target data labels, but they require large amounts of target training samples, which may not be available always. Also, semantic alignment of classes is difficult without a shared feature space which can be sought by creating positive and negative pairs using the source and target data \cite{Unified-SDA,Factor-SLN-ICDM-2014,SQSS-CVPR-2016,Typicality-MM-2007,Flickr-Groups-IR}.

Our framework is closely related to the adversarial adaptive methods \cite{Sim-deep-transfer,ADDA,Adaptation-BPP} particularly in the employment of view confusion loss (\ie cross-entropy loss). However, these works on domain adaptation are in the case of unlabelled target domains, and their ultimate goal is to regulate the learning of the source and target mappings so as to minimise the distance between the empirical mapping distributions. They chose an adversarial loss to minimise domain shift, learning a representation that is simultaneously discriminative of source labels while being able to distinguish between domains. Our method is not designed to match the target distribution to the source through an adversarial loss. In stead, we allow individual mappings which are not enforced to have weight sharing or any consistency to characterise view shifts and the adaptation is achieved through a view-adversarial loss.

\begin{figure*}[hbt]
\centering
\includegraphics[width=16cm,height=4cm]{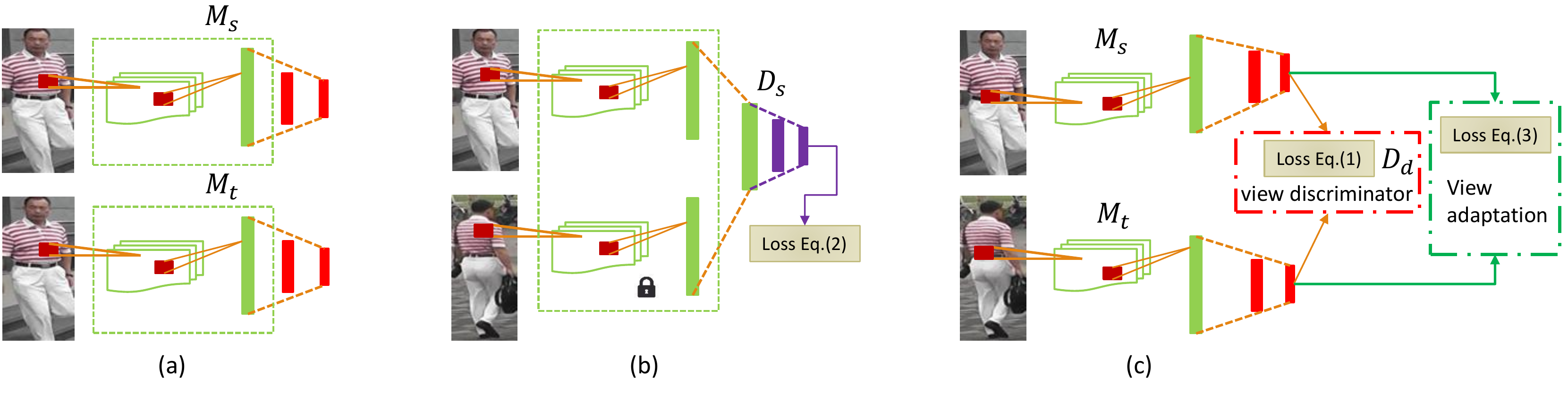}
\caption{The overview of adversarial adaptation learning for cross-view person re-ID. The network is trained with stage-wise updating on parameters (See text for details). Stage (a): The two mappings $\{M_s,M_t\}$ corresponding to different views are initialised by two CNNs. Stage (b): The similarity discriminator $D_s$ is trained by optimising the loss function Eq.\eqref{eq:sim-discriminator}. Stage (c): The view discriminator $D_d$ is trained by optimising the Eq.\eqref{eq:domain-discriminator}, and the mappings are updated by minimising the Eq.\eqref{eq:mapping-adversarial} to achieve view adaptation.}\label{fig:adversarial-training}
\end{figure*}

\section{Our Approach}\label{sec:approach}

\subsection{Problem Formulation}
We consider the training task where $\{X_s, X_t\}$ is the input space with $X_s$ and $X_t$ containing person images captured by two disjoint cameras, namely probe view (source) and gallery view (target). Specifically, the model is trained on labeled pairs in correspondence $(x_s^i, x_t^i)$ where $x_s^i$ and $x_t^i$ are examples of the same person $i$ across camera views. To address the view variance, we formulate it into adversarial adaptive manner: the main goal is to regularise the learning of the source and target mappings, $M_s$ and $M_t$, so as to minimise the distance between the empirical source and target mapping distributions: $M_s(x_s^i)$ and $M_t(x_t^i)$. Under this setting, the similarity discriminator is to learn to directly determine the input pair $(x_s^i, x_t^i)$ belongs to the same person or not, eliminating the cross-view variance.

The standard generative adversarial learning pits two networks against each other: a discriminator and a generator. The generator is in principle trained to produce images in a way that confuses the discriminator, which in turn tries to distinguish them from real image examples. In our case of cross-view adaptation for matching persons, this principle is employed to ensure that the networks cannot distinguish between the distribution of its probe view ($X_s$) and gallery view ($X_t$) \cite{CoGAN,Sim-deep-transfer,Adaptation-BPP}. In other words, a view discriminator ($D_d$) is adopted to classify whether an example is from the source or the target view. However, the generator is not needed in our network because generative modelling of input image distributions is not necessary, as the ultimate task is to learn discriminative representations regarding identities. On the other hand, asymmetric mappings can better model the differences of camera views than symmetric ones. Therefore, we first learn a couple of asymmetric mappings conditioned on each other through the view-adversarial training to produce view-invariant feature space. Then, a similarity estimator ($D_s$) with a margin-based separability is optimised on the Euclidean distances of positive/negative pairs to learn the effective similarity metrics.

\subsection{Discriminator Networks}

In our full adversarial adaptation framework, we have a \textbf{view discriminator} $D_d$, which classifies whether a data point is drawn from the probe or the gallery domain. Thus, $D_d$ can be optimised according to a supervised loss, and the label indicates the origin domain. Herein, $D_d$ is defined below:
\begin{equation}\label{eq:domain-discriminator}
\begin{split}
& \min_{D_d} \mathcal{L}_{D_d}(X_s, X_t, M_s, M_t) =-E_{x_s\sim X_s}\left[ \log D_d (M_s (x_s) ) \right]\\
& - E_{x_t\sim X_t}\left[ \log (1- D_d (M_t (x_t))) \right],
\end{split}
\end{equation}
where we design the individual probe and gallery mappings $M_s$ and $M_t$. It is clear that the two mappings are both parameterised in the supervised training with their asymmetric structures. This strategy is different from existing discriminative domain adaptation approaches \cite{ADDA} which generally consider a separate adaptation: the probe mapping is first learned through supervised losses, and then target mappings are initialised while adapting with the probe. By contrast, we aim to ensure the distance minimisation between the probe and gallery domains under their respective mappings, while crucially maintaining both mappings semantically discriminative. To effectively minimise the view discrepancy, we design the view-adversarial mapping loss (as defined in Eq.\eqref{eq:mapping-adversarial}) which suits the case where we initially use independent mappings  and then the galley mapping ($M_t$) is updated to adversarially to match the probe ($M_s$).

An effective re-ID system requires a metric to estimate the similarity for the paired pedestrian snapshots. This is amount to learning discriminative representations that are able to distinguish positive pairs against negative ones. Thus, to empower the view-adaptation framework with discriminative capability, we propose to optimise a view-invariant feature space such that data examples with the same identity are closer than those with different identities. In this work, we are interested in performing an end-to-end training for each paired images in their RGB values and optimising the view-discrepancy jointly with their similarity metrics. As a result, we can simply estimate similarity values for persons by directly computing the Euclidean distances of their embeddings.

\begin{figure}[hbt]
\centering
\includegraphics[width=9cm,height=3.5cm]{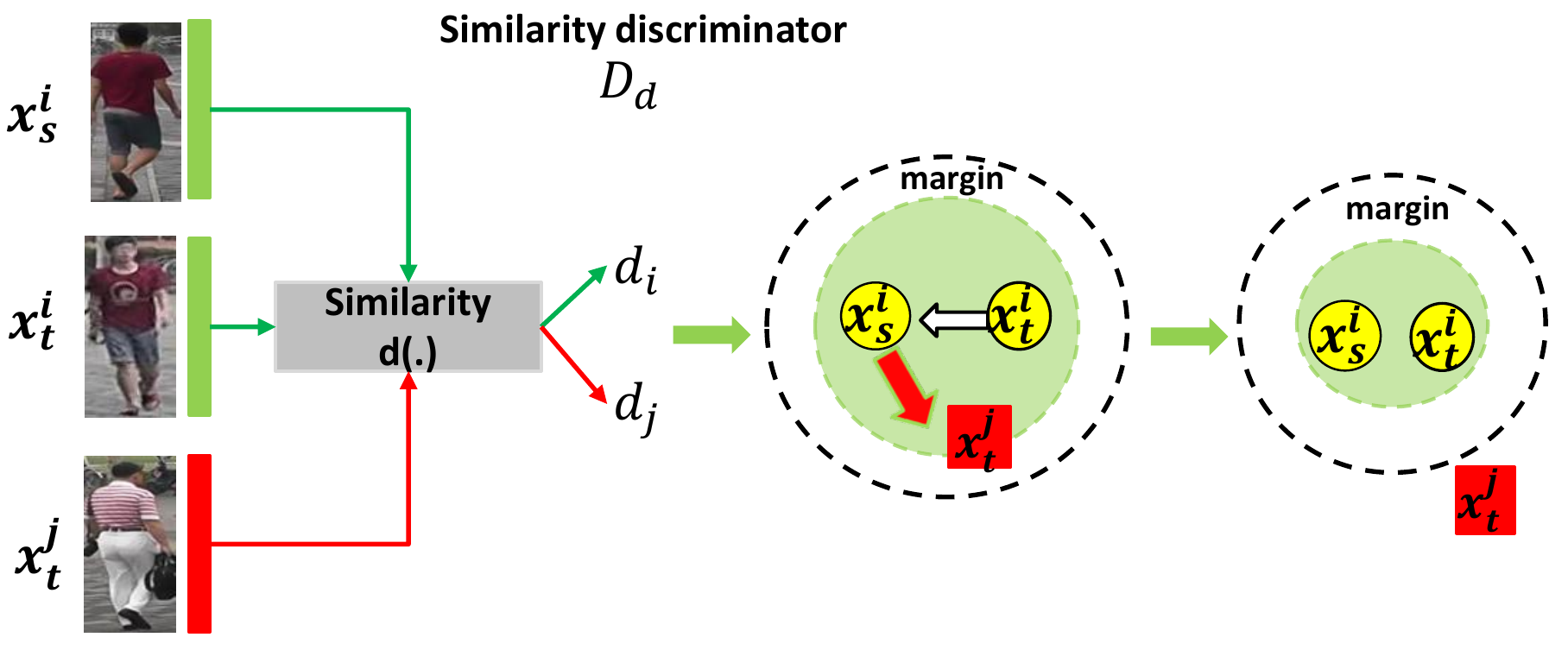}
\caption{The similarity discriminator learns to embed images into their Euclidean distances ($d(\cdot)$) whereby the loss is to minimise the distances between positive examples, and maximise the distances between negative ones. This is motivated by the nearest-neigh classification by enforcing a margin between positive and negative pairs.}\label{fig:discriminator}
\end{figure}

To generate the embedding for each pair $(x_s,x_t)$ \footnote{We omit the superscript $i$ for the notation simplification.}, and the corresponding similarity metric, we adopt the \textbf{similarity discriminator} network $D_s(\cdot)$ that aims to map semantically similar examples onto metrically close while simultaneously map semantically different examples onto metrically distant points in the embedding space. Hence, we formulate the similarity discriminator to minimise the following loss:
\begin{equation}\label{eq:sim-discriminator}\small
\begin{split}
& \min_{M_s,M_t, D_s} \mathcal{L}_{D_s}(X_s, X_t, Y) =\\
& -E_{(x_s, x_t)\sim(X_s, X_t)} \left[ \sum_{(x_s, x_t)} y \log D_s (M_s(x_s), M_t(x_t)) \right]\\
& - \gamma E_{(x_s, x_t)\sim(X_s, X_t)} \left[ \sum_{(x_s, x_t)} (1- y ) \{ \max(0, m-d(x_s, x_t))\}^2 + y d(x_s, x_t)^2 \right],
\end{split}
\end{equation}
where $y\in Y$ is the binary label assigned to the pair $(x_s, x_t)$, and $y=1$ if the pair is positive and $y=0$ otherwise. $Y$ denotes the number of identities in training. $d(x_s, x_t)$ denotes the Euclidean distance between two input vectors: $d(x_s, x_t)=|| M_s(x_s)-M_t(x_t)||_2$. $m$ is the margin that defines the separability in the embedding space and $\gamma$ is the parameter to control the relative importance of two losses. In our experiments, $m$ is empirically set to be $m=2$ and $\gamma$ is set to be $\gamma=2.5$ (see the empirical evaluations in Section \ref{sec:exp}). The scheme of similarity discriminator is illustrated in Fig.\ref{fig:discriminator}.

\subsection{Adversarial View Adaptation with Cross-Entropy Loss}

In our framework, we need to minimise the distances between the probe and gallery representations through alternating the minimisation between two functions. Thereby, the probe and gallery mappings should be optimised according to a constrained adversarial objective \cite{Sim-deep-transfer}, which can be formulated as:
\begin{equation}\label{eq:mapping-adversarial}\small
\begin{split}
&  \min_{M_s,M_t} \mathcal{L}_{M}(X_s, X_t, D_d) = \\
& -\sum_{d\in \{s,t\} } E_{x_d \sim X_d} \left[ \frac{1}{2} \log D_d (M_d (x_d) ) + \frac{1}{2} \log (1-D_d (M_d (x_d)) ) \right].
\end{split}
\end{equation}

Intuitively, the loss function in Eq.\eqref{eq:mapping-adversarial} is a view confusion objective, under which the mapping can be trained using a \textbf{cross-entropy} loss function against a uniform distribution. This loss is to ensure the adversarial discriminator will view the two domains identically. Finally, the full objective function is formulated to be the unconstrained optimisation as follow:
\begin{equation}\label{eq:full-obj}
\begin{split}
& \min_{D_d} \mathcal{L}_{D_d}(X_s, X_t, M_s, M_t); \\
& \min_{M_s,M_t} \mathcal{L}_{M}(X_s, X_t, D_d);\\
& \min_{M_s,M_t,D_s} \mathcal{L}_{D_s}(X_s,X_t,Y).
\end{split}
\end{equation}

The components of the objective function Eq.\eqref{eq:full-obj} can be interpreted as:

\begin{itemize}
\item $\min_{D_d} \mathcal{L}_{D_d}(X_s, X_t, M_s, M_t)$: We allow independent view mappings without enforcing weight sharing ($M_s \neq M_t$). This introduces a more flexible learning paradigm that allows view specific feature extractions to be learned. Siamese-like networks in person re-ID \cite{What-and-where,SI-CI,Multi-loss} have layer-wise equality constraint, thus enforcing exact probe and target mapping consistency. Indeed, learning a symmetric transformation reduces the number of parameters in the model, and ensures the mapping is view-invariant when the optimisation is converged. However, this may render the optimisation poorly conditioned because the same network is demanded to deal with images from two separate distributions.

\item $\min_{M_s,M_t} \mathcal{L}_{M}(X_s, X_t, D_d)$: In the setting where both the mappings are changing, the standard GAN loss cannot be applied because in the GAN setting the source distribution remains fixed while the target distribution is learned to match it. Thus, we aim to optimise the view confusion objective, and the mappings are updated using cross-entropy loss against a uniform distribution.

\item $\min_{M_s,M_t,D_s} \mathcal{L}_{D_s}(X_s,X_t,Y)$: We choose $\mathcal{L}_{D_s}(\cdot)$ to be a discriminative base model, as most prior adversarial adaptive methods suggest a generative model is not necessary while optimisation can be performed directly in a discriminative space for this purpose \cite{Few-shot-DA,ADDA}.
\end{itemize}

\begin{algorithm}[t]
\KwIn{Paired person images in cross-view $\{X_s,X_t\}$ where pairs are labeled in correspondence.}
\KwOut{A similarity discriminator $D_s$ and cross-view mappings $M_s$ and $M_t$.}
Initialise two mappings $M_s$ and $M_t$ using M-Net and D-Net.

Initialise $D_s$ using VGG pre-trained on ImageNet and fine-tuned on a soft-max function.

Uniformly sample $(x_s, x_t)$ from $\{X_s,X_t\}$.

Train $D_s$ using Eq.\eqref{eq:sim-discriminator}.

\While{not convergent} {
    Train $D_d$ by minimising Eq.\eqref{eq:domain-discriminator}.

    Update $M_s,M_t$ by minimising Eq.\eqref{eq:mapping-adversarial}.
}
\Return{$D_s$, $M_s,M_t$.}
\caption{{\sc Cross-Entropy} adversarial view adaptation learning for person re-ID.}
\label{algo:train-ANNs}
\end{algorithm}

\subsection{Network Training}
We optimise the objective function Eq.\eqref{eq:full-obj} in stages. The overall network has three components to be trained: a similarity discriminator $D_s$, a view discriminator $D_d$, and mappings across views $\{M_s,M_t\}$.  First, $M_s$ and $M_t$ are initialised by two deep models: M-Net \cite{M-Net} and D-Net \cite{VGG}, which are effective in independent feature detection and extraction \cite{What-and-where,Deep-recursive}. Then, the similarity discriminator $D_s$ is modelled by stacked fully-connected layers: 1024 hidden units, 2048 hidden units, and the final similarity output. With the exception of the similarity output layer, these fully-connected layers are using a ReLU activation function. However, there would be a severe imbalance between the number of within-identity pairs and the much greater number of between-identity pairs because the model requires the access to all pairs as input. Thus, our first improvement is to introduce an adaptive weighted loss into the similarity discriminator for the sake of imbalance.

\subsubsection{Adaptive Weighted Loss}

The challenge of learning effective features during training with a balanced model is to assign larger weights to difficult positive and negative samples \cite{MTMCT}. We improve the similarity loss in Eq. \eqref{eq:sim-discriminator} by introducing adaptive weight distribution on the positive/negative class. Thus, Eq.\eqref{eq:sim-discriminator} can be rewritten as:
\begin{equation}\label{eq:weighted-loss}\small
\begin{split}
& \min  \mathcal{L}_{D_s}^{\ast} = \sum_{x_p\in P(x_s)} \left[ \log D_s (M_s(x_s), M_t(x_p)) + w_p d(x_s, x_p)^2 \right]\\
& - \gamma \sum_{x_n \in N(x_s)} \left[ \{ \max(0, m- w_n d(x_s, x_n))\}^2 \right],
\end{split}
\end{equation}
where the gallery sample $x_t\in X_t$ is positive to $x_s$, i.e. $x_p\in P(x_s)$ or negative to $x_s$, i.e., $x_n\in N(x_s)$. $w_p$ and $w_n$ denote the weights assigned to the positive and negative pairs, respectively.Through this adaptive weight loss, the positive/negative class imbalance is alleviated by the explicit reflection on weight distribution. Apparently, the advantage of this adaptive weighing on positive/negative samples is to pertain the contribution of hard samples whilst the original loss using the uniform weights can eliminate the effect of hard samples, and thus very likely to get into the local minima as driven by easy samples. In our implementation, $w_p$ and $w_n$ are defined by using the soft-max/min weight distributions as:
\begin{equation}\label{eq:weight-definition}
w_p=\frac{\exp^{d(x_s,x_p)}}{\sum_{x_p\in P(x_s)} \exp^{d(x_s,x_p)}}; w_n=\frac{\exp^{-d(x_s,x_n)}}{\sum_{x_n\in N(x_s)} \exp^{d(x_s,x_n)}}.
\end{equation}

In the training, the VGG-16 network \cite{VGG} pre-trained on ImageNet \cite{ILSVRC} is used as the base feature architecture. Following the conventional fine-tuning strategy \cite{LSRO}, the last fully-connected layer is modified to have $K$ neuron to predict the $K$-classes, where $K$ is the number of training persons. Once fine-tuning is done, the convolutional layers of VGG architecture are used to be the non-linear transformations for the two mappings. As the network takes paired inputs, the two mappings are not applied with weight-sharing to ensure the network asymmetric. The outputs of each pair is concatenated before passing into the similarity discriminator \cite{Conditional-GAN}. Once the $D_s$ and $\{M_s, M_t\}$ are trained, the next step is training the view discriminator $D_d$ by classifying the images into $X_s$ or $X_t$. We model $D_d$ by using two fully-connected layers with a soft-max activation in the last layer to optimise the loss function of Eq. \eqref{eq:domain-discriminator}. This is implemented by freezing the $D_s$, $\{M_s, M_t\}$ and updating the parameters of $D_d$. Then, the network is trained to confuse $D_d$ in which the cross-entropy loss is computed by optimising Eq. \eqref{eq:mapping-adversarial}. The training process is illustrated in Fig.\ref{fig:adversarial-training}, and the procedure is summarised in Algorithm \ref{algo:train-ANNs}.

To address the imbalance on training pairs, we improve the optimisation on the similarity discriminator $D_s$ by introducing the adaptive weighted loss. Thus, to construct a batch during training and calculate adaptive weights, we follow MTMCT \cite{MTMCT} to construct $\mathcal{S}\mathcal{P}$ batches. In specific, during a training epoch each identity is selected into its batch, and the remaining $\mathcal{P}-1$ batch identities are selected at random. And $\mathcal{S}$ samples for each identity are also selected at random.

\section{Experiments}\label{sec:exp}

\begin{figure}[t]
\centering
\includegraphics[width=9cm,height=3cm]{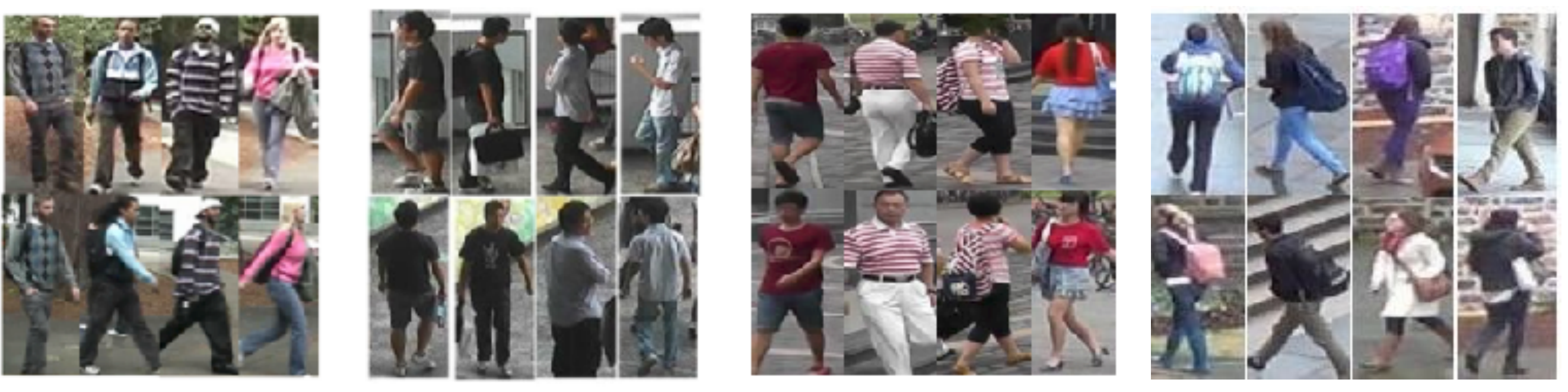}
\caption{Examples from person re-ID datasets. From left to right: VIPeR, CUHK03, Market-1501, and DukeMTMC-reID. Columns indicate the same identities.}\label{fig:person-datasets}
\end{figure}

\subsection{Datasets and Evaluation}
We perform extensive experiments and comparative studies to evaluate our approach over four benchmark datasets: VIPeR \cite{Gray2007Evaluating}, CUHK03 \cite{FPNN}, Market-1501 \cite{Market1501}, and DukeMTMC-reID \cite{DukeMTMC,LSRO}. Example images are shown in Fig.\ref{fig:person-datasets}.

\begin{itemize}
\item The \textbf{VIPeR} dataset \cite{Gray2007Evaluating} contains $632$ individuals taken from two cameras with arbitrary viewpoints and varying illumination conditions. The 632 person images are randomly divided into two equal halves, one for training and the other for testing.

\item The \textbf{CUHK03} dataset \cite{FPNN} includes 13,164 images of 1360 pedestrians. The whole dataset is captured with six surveillance camera. Each identity is observed by two disjoint camera views, yielding an average 4.8 images in each view. This dataset provides both manually labeled and detected pedestrian bounding boxes. Our experiments report results on the labeled dataset.

\item The \textbf{Market-1501} dataset \cite{Market1501} contains 32,668 fully annotated boxes of 1501 pedestrians. Each identity is captured by at most six cameras and boxes of person are obtained by running a state-of-the-art detector, the Deformable Part Model (DPM) \cite{MarketDetector}. The dataset is randomly divided into training and testing sets, containing 750 and 751 identities, respectively.

\item The \textbf{DukeMTMC-reID} dataset is a re-ID version of the DukeMTMC dataset \cite{DukeMTMC}. It contains 34,183 image boxes of 1,404 identities of which 702 are used for training and the remaining 702 for testing. The probe and gallery images are 2,228 and 17,661, respectively.
\end{itemize}

We evaluate all the approaches with Cumulative Matching Characteristic (CMC) results by the single-shot setting. The CMC curve can characterise a ranking result for every image in the gallery given the probe image. We also use mean Average Precision (mAP) as performance measure on CUHK03, Market-1501, and DukeMTMC-reID.

\subsection{Settings}

Considering the training of our network is accessible to few examples from each person because we do not perform data augmentation, it is necessary to perform cross-validation on hyper-parameters to improve the generalisation on unseen observations. We therefore construct two disjoint sets of classes to be $C_{train}$ and $C_{valid}$ on each train set of each dataset. For example, on VIPeR dataset, the three subsets are randomly divided to be $C_{train}$ (216 persons), $C_{valid}$ (100 persons), and $C_{test}$ (316 persons). The details of the train/validation/test division on four datasets are given in Table \ref{tab:set-division}. There are up to six cameras for CUHK03 and Market-1501, and thus for each person we randomly select two cameras to be the probe and gallery views. Then, each person's images across views are selected to be samples in pairs.

\begin{table}[t]
\caption{The division on train/validation/test set of each dataset.}\label{tab:set-division}
\centering
 \begin{tabular}{|c |c |c |c|}
 \hline
 Dataset & $\sharp C_{train}$  & $\sharp C_{valid}$  & $\sharp C_{test}$  \\
 \hline
 VIPeR & 216 & 100 & 316 \\
 \hline
 CUHK03 & 1160 & 100 & 100 \\
 \hline
 Market-1501 & 650 & 100 & 751 \\
 \hline
 DukeMTMC-reID & 602 & 100 & 702\\
 \hline
\end{tabular}
\end{table}

We use the VGG architecture and its variants M-Net and D-Net as the feature bases which are initialised from weights pre-trained on ImageNet and fine-tuned on target $C_{train}$ of each re-ID dataset. Once fine-tuning is done, the convolution layers of each network are used as $M_s$ ($M_t$), and a three-layer fully connection with ReLU as activation function is used as the similarity discriminator $D_s$. The hidden layers in $D_s$ have the dimensionality of 1,024 and 2,048, respectively. The learning rate starts with 0.001 and is divided by 10 every 10 epoches. The network uses a batch size of 128 images. The training is stopped when the loss stops decreasing during the validation on $C_{valid}$.

\subsection{Experimental Results}

\begin{figure*}[t]
\centering
\begin{tabular}{cc}
\includegraphics[width=7cm,height=5cm]{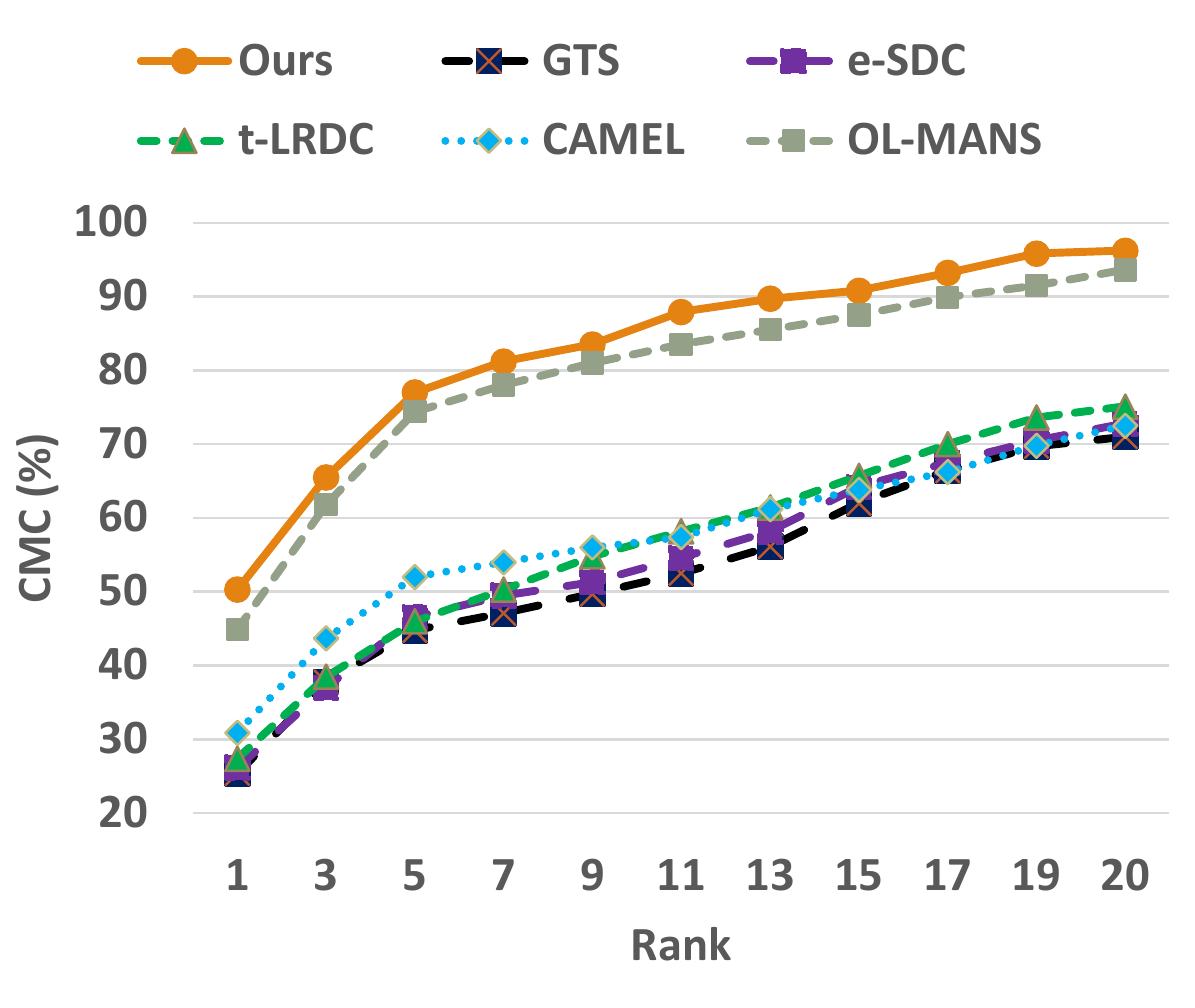} &
\includegraphics[width=7cm,height=5cm]{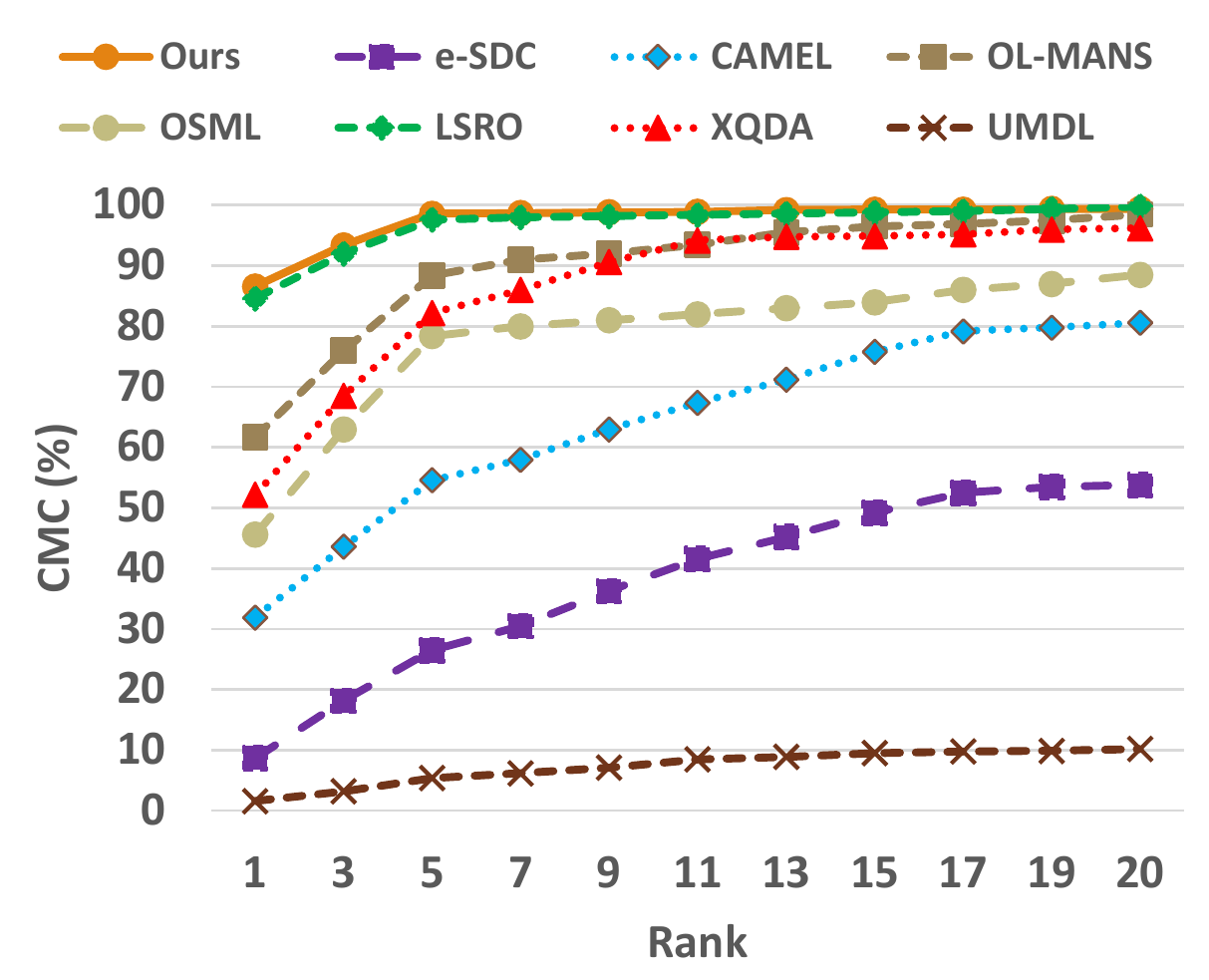} \\
(a) VIPeR & (b) CUHK03 \\
\includegraphics[width=7cm,height=5cm]{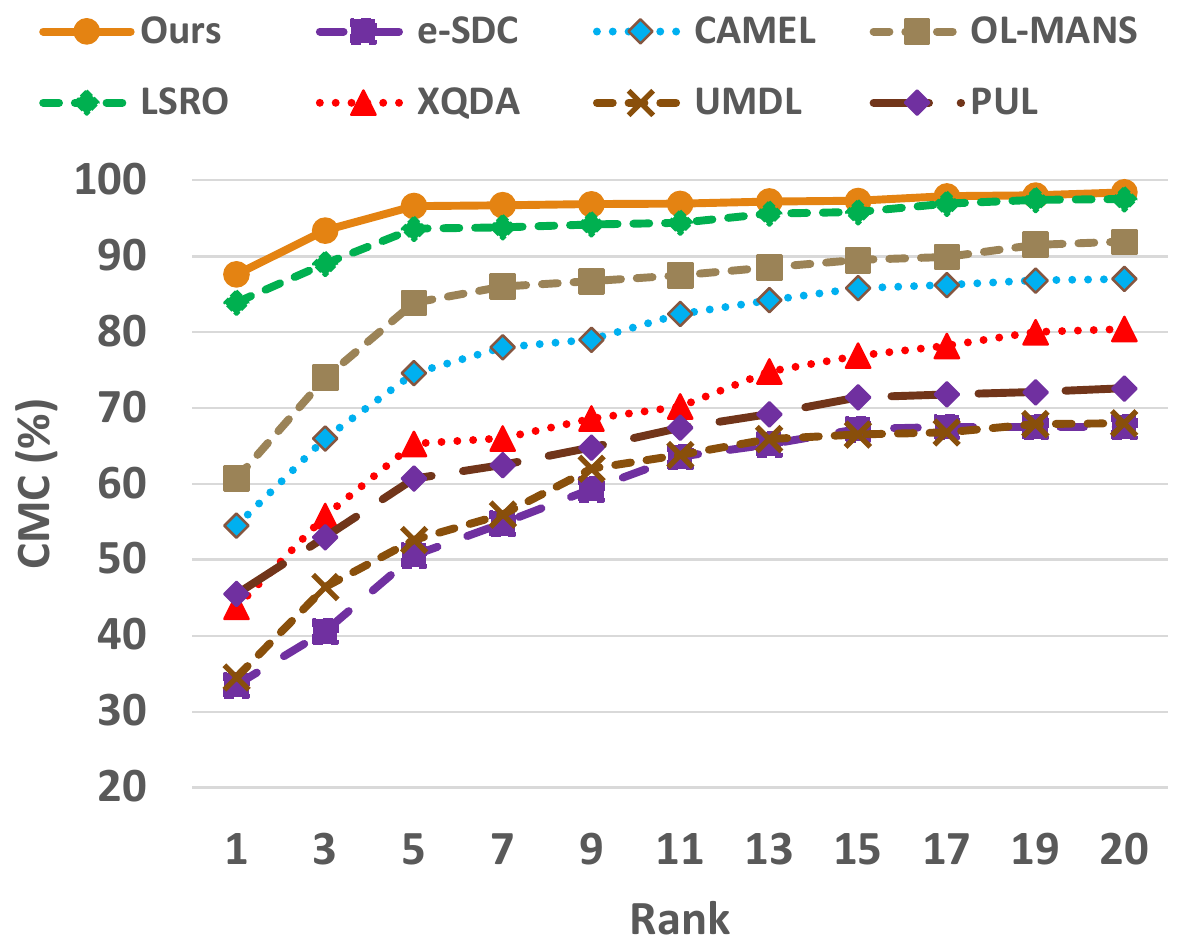}\ &
\includegraphics[width=7cm,height=5cm]{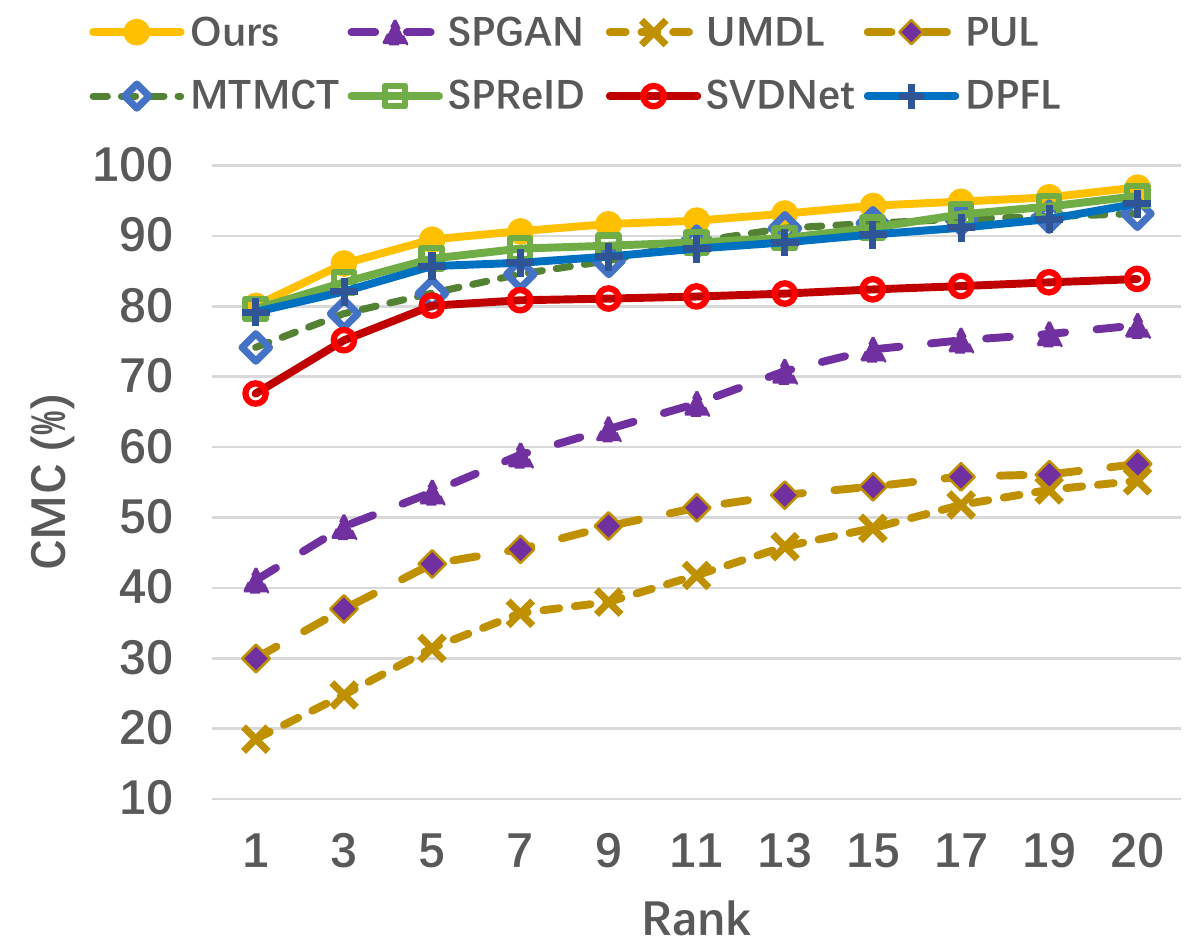}\\
(c) Market-1501 & (d) DukeMTMC-reID
\end{tabular}
\caption{CMC curves of different methods on the four datasets.}\label{fig:cmc}
\end{figure*}

In this section, we compared the proposed method with recent un/semi-supervised and supervised models on four datasets. The comparison results measured by rank-$R$ accuracies of CMC are shown in Fig.\ref{fig:cmc}. And respective rank-$R$ ($R=1,5,20$) values on four datasets are given in Table \ref{tab:cmc-viper}, Table \ref{tab:cmc-cuhk03}, Table \ref{tab:cmc-market1501}, and Table \ref{tab:cmc-DukeMTMC-reID}. We also conduct self-ablation evaluations on parameter sensitivity and network architecture.

\paragraph{Comparison to Un/semi-supervised Methods}
We compared our method with several unsupervised re-ID models, including local salience learning based models (GST \cite{GenerativeSaliency} and eSDC \cite{eSDC}), transfer-learning based models (t-LRDC \cite{t-LRDC}, PUL \cite{PUL}, and UMDL \cite{UMDL}), metric learning methods (OSML \cite{One-shot-RE-ID}, CAMEL \cite{CAMEL}, OL-MANS \cite{OL-MANS}), and a semi-supervised method of LSRO \cite{LSRO}.

On the VIPeR dataset, Table \ref{tab:cmc-viper} shows that our method outperforms other models in the case when there is only one example for each person in each view. For example, our method achieves rank-1=51.3, which is noticeably improved performance compared to OL-MANS \cite{OL-MANS} with rank-1=44.9. The main reason is that the assumptions without supervision cannot provide the view-specific inference, and thus impedes these unsupervised methods from achieving higher accuracies. In contrast, the proposed method is based on adversarial learning which is able to effectively minimise the view discrepancy without requiring large numbers of labeled training examples. Moreover, the improved variant of our approach (denoted as $Ours^{\ast}$) with adaptive weighted loss can emphasise the most difficult samples in a batch, an thus outperforms the state-of-the-art SpindleNet \cite{SpindleNet} at rank-1 value.

On the CUHK03 dataset, in Table \ref{tab:cmc-cuhk03} it can be seen that our method outperforms the state-of-the-art by large margins. For instance, the rank-1 value is improved by 25\% compared to OL-MANS \cite{OL-MANS}. The reality is the illumination changes in CUHK03 are extremely severe and even human beings may find difficulty in identifying the persons across views. Without the aid of supervision, unsupervised methods cannot retrain the appearance robustness against visual variations. As a comparison, our approach is able to address this issue by training a discriminative distance metric jointly with the view-invariant feature learning. This leads to better performance of the proposed method. Also, the performance of our method with adaptive weighting outperforms the state-of-the-art SpindleNet \cite{SpindleNet} which builds the discriminative representations by extensive body region decompositions.

Table \ref{tab:cmc-market1501} and Table \ref{tab:cmc-DukeMTMC-reID} report the comparison results on the Market-1501 and DukeMTMC-reID datasets, respectively. Our method has achieved notable performance gain on the two datasets in comparison with these un/semi-supervised methods. These empirical evaluations on different benchmark datasets demonstrate the effectiveness of our model in cross-view person re-ID owing to the effective view adaptation while learning discriminative metrics in the context of view-aligned feature space.

\begin{table}[t]
  \centering
  \begin{tabular}{|l|l|c|c|c|}
    \hline
     \multicolumn{2}{|c|}{\cellcolor{gray} \textcolor{white} {Method}} & \cellcolor{gray} \textcolor{white} {R=1} & \cellcolor{gray} \textcolor{white} {R=5} & \cellcolor{gray} \textcolor{white} {R=20} \\
     \hline
     & Ours & 51.3 & 77.0 & 96.1 \\
     & $Ours^{\ast}$ & $\mathbf{55.9}$ & $\mathbf{79.2}$ & $\mathbf{97.9}$ \\
        \hline
   \parbox[t]{1mm}{\multirow{6}{*}{\rotatebox[origin=c]{90}{Un/s-supervised}}} & GTS \cite{GenerativeSaliency} & 25.2 & 44.8 & 71.0\\
   & eSDC \cite{eSDC} & 26.3 & 46.6  & 72.8 \\
   & t-LRDC \cite{t-LRDC} & 27.4 & 46.0 & 75.1\\
   & OSML \cite{One-shot-RE-ID} & 34.3 & -& -\\
   & CAMEL \cite{CAMEL} & 30.9 & 52.0 & 72.5\\
   & OL-MANS \cite{OL-MANS} & 44.9 & 74.4& 93.6\\
    \hline
   \parbox[t]{1mm}{\multirow{8}{*}{\rotatebox[origin=c]{90}{Supervised}}} & DCSL \cite{DCSL} & 44.6 & 73.4 & 82.6\\
   & JSTL \cite{DG-Dropout} & 20.9  & - & -\\
   & Deep-Embed \cite{Deep-Embed} & 49.0  & 77.1  &96.2 \\
   & SpindleNet \cite{SpindleNet} & 53.8 & 74.1 & 92.1\\
   & Part-Aligned \cite{Part-Aligned} & 48.7 & 74.7 & 93.0\\
   & DNSL \cite{NullSpace-Reid} & 42.3 & 71.5  & 92.1 \\
   & SI-CI \cite{SI-CI} & 35.8 & 72.3  & 97.1\\
   & PIE \cite{PIE-reid} & 18.1& 25.3 & 49.4\\
    \hline
  \end{tabular}
  \caption{Comparison results with state-of-the-arts on VIPeR. Un/s-supervised stands for unsupervised/semi-supervised. The best results are in bold.}\label{tab:cmc-viper}
\end{table}

\paragraph{Comparison to Supervised Methods}
We compared the proposed method against recent state-of-the-art supervised models: DCSL \cite{DCSL},  JSTL \cite{DG-Dropout}, DNSL \cite{NullSpace-Reid}, Deep-Embed \cite{Deep-Embed}, SpindleNet \cite{SpindleNet}, Part-Aligned \cite{Part-Aligned}, MSCAN \cite{MSCAN}, SI-CI \cite{SI-CI}, PIE \cite{PIE-reid}, JLML \cite{Multi-loss}, MTMCT \cite{MTMCT},  SPReID \cite{SPReID}, SVDNet \cite{SVDNet}, and DPFL \cite{DPFL}. Comparison results on three datasets are reported in Table \ref{tab:cmc-viper}, Table \ref{tab:cmc-cuhk03}, Table \ref{tab:cmc-market1501},  and Table \ref{tab:cmc-DukeMTMC-reID} respectively. It can be noticed that our method achieves better results compared to these supervised methods, and can outperform them when the adaptive weighting is applied. In Table \ref{tab:cmc-viper}, we obtain rank-1=51.3 (55.9 from $Ours^{\ast}$) on the VIPeR dataset which has gained the recognition improvement over the SpindleNet \cite{SpindleNet} by 2.1\% in rank-1 value. And in Table \ref{tab:cmc-cuhk03}, we obtain rank-1=86.6 (88.9 after weighted adaptation) as opposed to SpindleNet \cite{SpindleNet} with rank-1=88.5. We remark that SpindleNet \cite{SpindleNet} is a fully-supervised method that needs annotations on each body region to focus/extract these local features to describe each person. The process of annotating each body region is very cumbersome and not scalable in large networked cameras. On Market-1501, comparison results in Table \ref{tab:cmc-market1501} show that our method greatly improves the rank-1 accuracy for this task. For example, in comparison with MSCAN \cite{MSCAN}, a state-of-the-art method based on fully-supervised body region encoding, the rank-1 accuracy value goes from 80.3\% up to 87.2\%. This is particularly effective in Market-1501 dataset where each person has up to 10 samples, and our approach is able to address the view misalignment more carefully. Our approach with weight adaptation loss ($Ours^{\ast}$) can further improve the rank-1 accuracy and achieve 89.1\%, which is better than the state-of-the-art SPReID* (rank-1=88.3\%)\cite{SPReID} \footnote{Please note that all results of SPReID \cite{SPReID} are reported by using  reduced data augmentation backboned on ResNet-152 architecture.} and DPFL \cite{DPFL} (rank-1=88.6\%). Experimental results on the DukeMTMC-reID dataset are reported in Table \ref{tab:cmc-DukeMTMC-reID}. Our method outperforms the state-of-the-art DPFL \cite{DPFL} by 1\% at rank-1 accuracy. It shows that the adaptive weighting scheme is very effective in training a balanced model on DukeMTMC-reID dataset which has severe imbalance classes in the probe (2,228 images) and gallery size (17,661 images).

\begin{table}[t]
  \centering
  \begin{tabular}{|l|l|c|c|c|c|}
    \hline
     \multicolumn{2}{|c|}{\cellcolor{gray} \textcolor{white} {Method}} & \cellcolor{gray} \textcolor{white} {R=1} & \cellcolor{gray} \textcolor{white} {R=5} & \cellcolor{gray} \textcolor{white} {R=20} & mAP\\
     \hline
     & Ours & 86.6 &  $\mathbf{98.6}$ & 99.4 & $\mathbf{91.4}$ \\
     & $Ours^{\ast}$ & $\mathbf{88.9}$ & $\mathbf{99.2}$ & $\mathbf{99.9}$ & $\mathbf{91.8}$ \\
    \hline
   \parbox[t]{1mm}{\multirow{7}{*}{\rotatebox[origin=c]{90}{Un/s-supervised}}}  & eSDC \cite{eSDC} & 8.7 & 26.5 & 53.4 & -\\
   & OSML \cite{One-shot-RE-ID} & 45.6 & 78.4& 88.5 &-\\
   & LSRO \cite{LSRO} & 84.6& 97.6 & 99.8 & 87.4\\
   & CAMEL \cite{CAMEL} & 31.9 & 54.6 & 80.6 & -\\
   & XQDA \cite{LOMOMetric} & 52.2 & 82.2 & 96.2 & 51.5\\
   & UMDL \cite{UMDL} & 1.6 & 5.4& 10.2 & -\\
   & OL-MANS \cite{OL-MANS} & 61.7 & 88.4 & 98.5 &-\\
    \hline
   \parbox[t]{1mm}{\multirow{9}{*}{\rotatebox[origin=c]{90}{Supervised}}} & DCSL \cite{DCSL} &80.2 &97.7& 99.8 &-\\
   & JSTL \cite{DG-Dropout} & 72.6 & 91.0 & 96.7 &-\\
   & Deep-Embed \cite{Deep-Embed} & 73.0 & 91.6 & 98.6 &-\\
   & SpindleNet \cite{SpindleNet} & 88.5 & 97.8 & 99.2 &-\\
   & Part-Aligned \cite{Part-Aligned} & 85.4 & 97.6 & 99.9 & 90.9\\
   & MSCAN \cite{MSCAN} & 74.2 & 94.3 & 99.3 & -\\
   & DNSL \cite{NullSpace-Reid} & 58.9 & 85.6  & 96.3 & -\\
   & SI-CI \cite{SI-CI} & 52.2 & 84.3  & 98.8 &-\\
   & PIE \cite{PIE-reid} & 62.4& 73.7& 95.6 & 71.3\\
   & SPReID \cite{SPReID}* & 88.0 & 95.2 & 99.9 &-\\
   & SVDNet \cite{SVDNet} & 68.5 & 90.2 & 94.0 & 73.3\\
   & DPFL \cite{DPFL} & 86.7 & 97.0 & 98.2 & 83.8\\
    \hline
  \end{tabular}
  \caption{Comparison results with state-of-the-arts on CUHK03. The best results are in bold.}\label{tab:cmc-cuhk03}
\end{table}

\begin{table}[t]
  \centering
  \begin{tabular}{|l|l|c|c|c|c|}
    \hline
     \multicolumn{2}{|c|}{\cellcolor{gray} \textcolor{white} {Method}} & \cellcolor{gray} \textcolor{white} {R=1} & \cellcolor{gray} \textcolor{white} {R=5} & \cellcolor{gray} \textcolor{white} {R=20} & mAP\\
     \hline
    & Ours & 87.2 &  96.3 & 98.5 & $\mathbf{74.7}$\\
    & $Ours^{\ast}$ & $\mathbf{89.1}$ & 96.8 & $\mathbf{99.7}$ & $\mathbf{76.2}$\\
    \hline
   \parbox[t]{1mm}{\multirow{7}{*}{\rotatebox[origin=c]{90}{Un/s-supervised}}} & eSDC \cite{eSDC} & 33.5 & 50.6 & 67.5 & 13.5\\
   & LSRO \cite{LSRO} & 83.9& 93.6 & 97.5 & 66.1\\
   & CAMEL \cite{CAMEL} & 54.5 &  74.6 & 87.0 &-\\
   & OL-MANS \cite{OL-MANS} &60.7 & 83.8& 91.9 &-\\
   & PUL \cite{PUL} & 45.5 & 60.7 & 72.6 &-\\
   & UMDL \cite{UMDL} & 34.5 & 52.6& 68.0  &-\\
   & XQDA \cite{LOMOMetric} & 43.8 & 65.3 & 80.4 &22.2\\
    \hline
   \parbox[t]{1mm}{\multirow{8}{*}{\rotatebox[origin=c]{90}{Supervised}}} & JSTL \cite{DG-Dropout} & 44.7 & 67.2 & 82.0 &-\\
   & Deep-Embed \cite{Deep-Embed} & 68.3 & 87.2  & 96.7 & 40.2\\
   & SpindleNet \cite{SpindleNet} & 76.9 & 91.5 & 96.7 &-\\
   & Part-Aligned \cite{Part-Aligned} & 81.0 & 92.3 & 97.1 &-\\
   & MSCAN \cite{MSCAN} & 80.3 & 92.0 & 97.0 & 57.5\\
   & DNSL \cite{NullSpace-Reid} & 55.4  & 75.0 & 87.3 & 35.7\\
   & PIE \cite{PIE-reid} & 79.3 & 90.7 & 96.5 & 55.9\\
   & JLML \cite{Multi-loss} & 85.1 & 97.9 & 99.5 & 65.5\\
   & MTMCT \cite{MTMCT} & 82.1 & 93.5 & 98.1 & 68.0\\
   & SPReID \cite{SPReID}* & 88.3 & 93.6 & 98.5 & 72.9\\
   & SVDNet \cite{SVDNet} & 80.5 & 91.7 & 93.7 & 62.1\\
   & DPFL \cite{DPFL} & 88.6 & 94.5 & 98.0  & 73.1\\
    \hline
  \end{tabular}
  \caption{Comparison results with state-of-the-arts on Market-1501. All results are evaluated on single-shot setting. The best results are in bold.}\label{tab:cmc-market1501}
\end{table}

\begin{table}[t]
  \centering
  \begin{tabular}{l|c|c|c|c}
    \hline
     Method & R=1 & R=5 & R=20 & mAP\\
     \hline
    Ours & 47.2 &  61.0 & 84.7 & 31.4\\
    $Ours^{\ast}$ & $\mathbf{80.1}$ & $\mathbf{89.5}$ & $\mathbf{96.9}$ & $\mathbf{67.2}$\\
    PUL \cite{PUL} & 30.0 & 43.4 & 57.6 & 16.4 \\
    UMDL \cite{UMDL} & 18.5 & 31.4 & 55.2 & 7.3 \\
    SPGAN \cite{SPGAN} & 41.1 & 56.6 & 77.3 & 22.3 \\
    MTMCT \cite{MTMCT} & 74.2 & 81.9 & 93.2 & 54.9\\
    SPReID \cite{SPReID}* & 79.6 & 86.8 & 95.7 & 62.4\\
    SVDNet \cite{SVDNet} & 67.6 & 80.5 & 83.7 & 45.8\\
    DPFL \cite{DPFL} & 79.2 & 85.7 & 94.6 & 60.6\\
    \hline
  \end{tabular}
  \caption{Comparison results with state-of-the-arts on DukeMTMC-reID. All results are evaluated on single-query setting. The best results are in bold.}\label{tab:cmc-DukeMTMC-reID}
\end{table}

\begin{figure}[t]
\centering
\begin{tabular}{c}
\includegraphics[width=8cm,height=5cm]{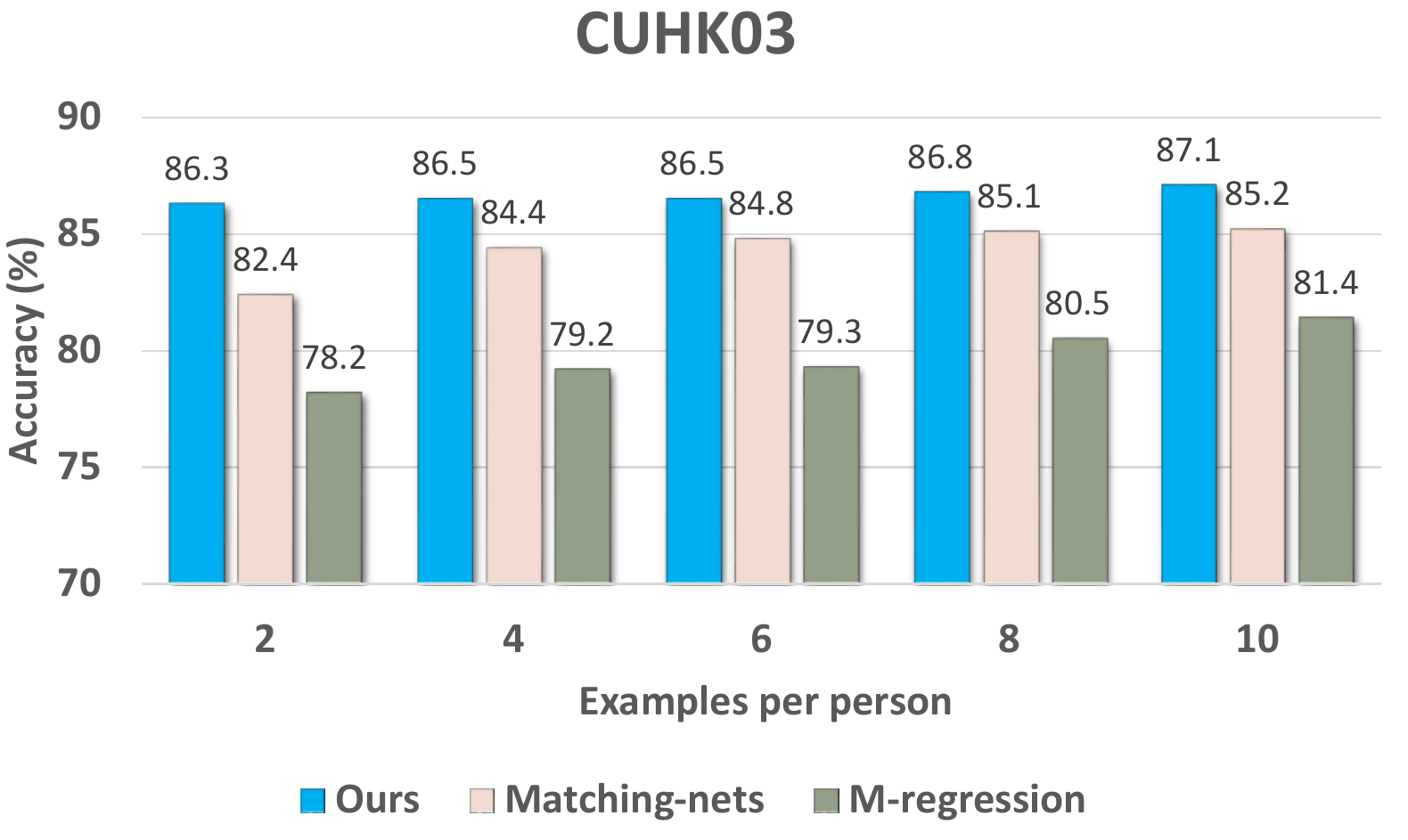} \\
\includegraphics[width=8cm,height=5cm]{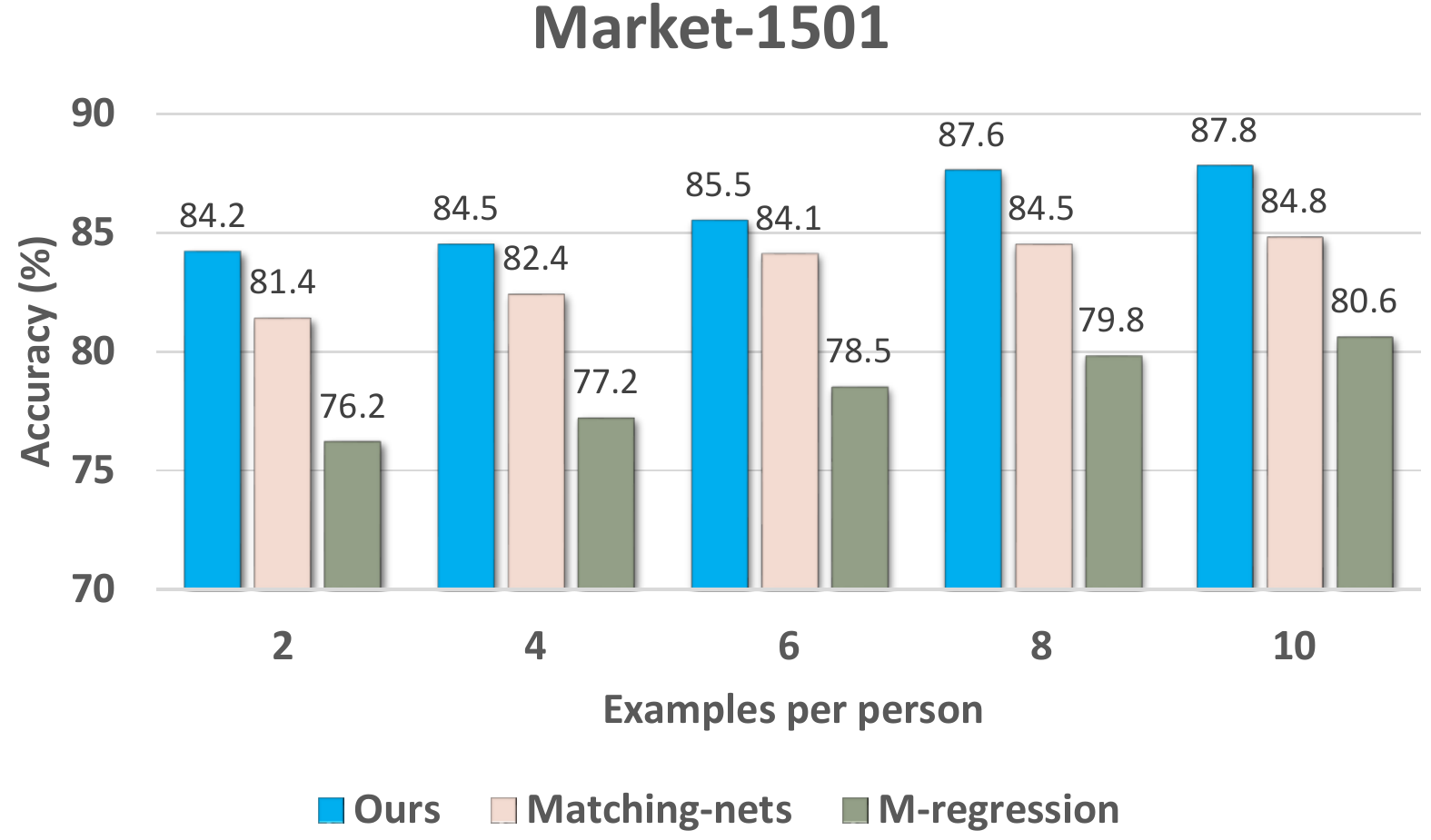} \\
\end{tabular}
\caption{Comparison to recent few-shot learning methods.}\label{fig:compare-few-shot}
\end{figure}

\paragraph{Self-Ablation Studies}

\begin{figure}[t]
\centering
\includegraphics[width=6cm,height=5cm]{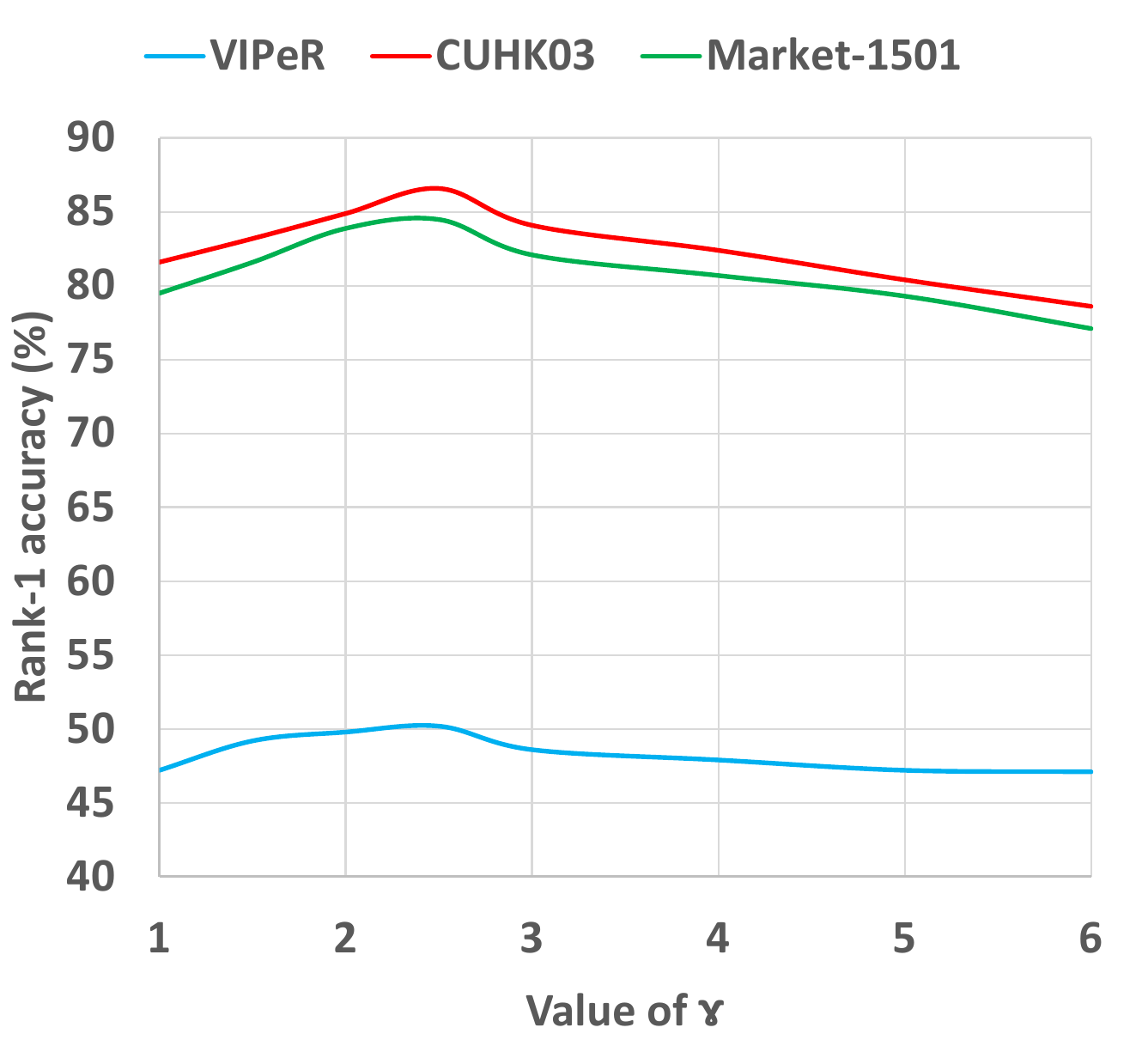}
\caption{$\gamma$ in Eq.\eqref{eq:sim-discriminator} w.r.t re-ID accuracy. A larger $\gamma$ indicates a larger weight of similarity discrimination constraint.}\label{fig:gamma}
\end{figure}

\begin{table}[t]
  \centering
  \begin{tabular}{c|c|c}
    \hline
     Architectures & VIPeR (R=1)& CUHK03 (R=1)\\
     \hline
    M-Net, D-Net = $M_s, M_t$ & 51.3 & 86.6 \\
    D-Net, M-Net = $M_s, M_t$ & 49.7 & 83.9 \\
    ResNet, ResNet = $M_s, M_t$ & 51.3 & 86.4 \\
    \hline
  \end{tabular}
  \caption{The study on different architectures.}\label{tab:study-architecture}
\end{table}

We first study the sensitivity of our model to the key parameter of $\gamma$ in Eq.\eqref{eq:sim-discriminator}. The impact of $\gamma$ is investigated and the results are shown in Fig.\ref{fig:gamma}. As $\gamma$ is to balance the relative importance of the discriminative distance metric, it is proven to have higher rank-1 accuracy when $\gamma=2.5$, while a larger $\gamma$ does not bring more gains in accuracy. Thus, we empirically set $\gamma=2.5$ in all experiments. We also study different network architectures to inspect the importance of backbone networks. In our experiment, we consider the VGG-16 and the ResNet \cite{ResNet}. Specifically, two variants of VGG: M-Net and D-Net are used to initialise ($M_s$, $M_t$), and two identical ResNet networks are employed to initialise ($M_s$, $M_t$) as a comparison. Experimental results are shown in Table \ref{tab:study-architecture}. We can observe that performances of two identical ResNet networks are inferior to the asymmetric architectures with M-Net and D-Net on VIPeR and CUHK03 datasets. Thus, we use M-Net and D-Net as default backbone networks.

\subsection{Comparison to Other Few-Shot Methods}
We also compared to two recently proposed few-shot learning methods: matching networks \cite{Match-nets} and model regression \cite{Model-regress}. The matching networks propose a nearest neighbour approach that trains an embedding end-to-end for the task of few-shot learning. Model regression trains a small MLP to regress from the classifier trained on a small dataset to the classifier trained on the full dataset. Both of the two techniques are high-capacity in learning from few examples and facilitates the recognition in the small sample size regime on a broad range of tasks, including domain adaptation and fine-grained recognition. Comparison results are shown in Fig. \ref{fig:compare-few-shot}. In terms of the overall performance, our method outperforms the two competitors constantly over the two datasets. Matching networks exhibit similar performance to our method, however, matching networks are based on nearest neighbours and use the entire training set in memory, and thus they are more expensive in testing time compared with our method and model regressors.

\begin{table}[t]
\centering
\begin{tabular}{c|c|c|c}
\hline
& VIPeR & CUHK03 & Market-1501\\
Method & \includegraphics[width=3mm, height=6mm]{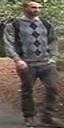}$\longleftrightarrow$ \includegraphics[width=3mm, height=6mm]{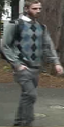} & \includegraphics[width=3mm, height=6mm]{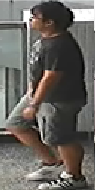}$\longleftrightarrow$ \includegraphics[width=3mm, height=6mm]{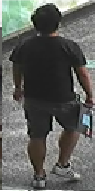}& \includegraphics[width=3mm, height=6mm]{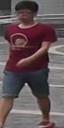}$\longleftrightarrow$ \includegraphics[width=3mm, height=6mm]{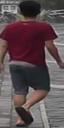}\\
\hline
SPGAN \cite{SPGAN} & -& -& 58.1\\
DAN \cite{DAN} & 39.6 & 71.0 & 53.8\\
CoGAN \cite{CoGAN} & 41.7 & - & 48.1\\
ADDA \cite{ADDA} & 39.4 & 73.1 & 56.4\\
Ours & $\mathbf{51.3}$ & $\mathbf{86.6}$ & $\mathbf{87.2}$\\
\hline
\end{tabular}
\caption{The comparison results with view adaptation methods.}\label{tab:compare-adaptation}
\end{table}

\subsection{Comparison with View Adaptation Methods}

In this experiment, we validate our approach in view adaptation by comparing to recent domain adaptation methods not limited from person re-ID: SPGAN \cite{SPGAN}, Deep Adaptation Networks (DAN) \cite{DAN}, Adversarial Discriminative Domain Adaptation (ADDA) \cite{ADDA}, and CoGANs \cite{CoGAN}. Experimental results are provided in Table \ref{tab:compare-adaptation}. For these domain adaptation methods including DAN \cite{DAN}, ADDA \cite{ADDA}, and CoGANs \cite{CoGAN}, their training are set and modified to adapt the gallery view (target) to match the probe view (source). For instance, CoGANs \cite{CoGAN} can learn a joint distribution of multiple-domain data, the learning can be conducted by using two generative models with an identical architecture corresponding to the probe and the gallery images of a person. Then, through weight sharing, CcGANs are able to encode high-level semantics regarding identities into the low-level feature extraction. Our approach achieves the highest rank-1 value on the three datasets, despite being trained without a deep generator yet being a considerably simpler model. This also provides compelling evidence that generating images is not necessarily relevant to effective view adaptation. This discovery is consistent with ADDA \cite{ADDA} which does not use a generative model while also shows convincing results in comparison with CoGANs \cite{CoGAN}. For CoGANs \cite{CoGAN}, it is sometimes hard to get convergence, e.g. on CUHK03 when the view changes are very disparate, and it is unable to train coupled generators for them simultaneously.

\section{Conclusion and Future Work}\label{sec:con}

In this paper, we introduce an effective view adaptation model to person re-identification to produce asymmetric transformations that can fully characterise view specificity. The approach is based on adversarial learning to minimise view-discrepancy through view confusion objective with an entropy regularisation to align and form view-invariant feature space. The network is trained with a cross-entropy loss to optimise view confusion objective and jointly with a discriminative distance metric through a margin-based separability criterial. Also, training imbalance is explicitly described as weight distribution on hard samples, and the proposed adaptive weighting loss can address it more effectively. Experimental results show that the adversarial neural networks are able to produce feature space with cross-view variations being reduced. The proposed approach works effectively for labeled training samples with large visual divergence, and our method shows clear promise as it sets new state-of-the-art performance in experiments.

In future work, we would explore the direction of view adaptation in the case when such training pairs are not given. One possibility is to learn a probe to gallery encoder-decoder under a generative adversarial objective with some reconstruction term which can be applied to predict the clothing people are wearing. The other interesting direction is towards intriguing few-shot learning principles to learn to match persons with more powerful augmented memory networks.

\section*{Acknowledgement}

Meng Wang was supported by he National Key Research and Development Program of China under grant 2018YFB0804200; NSFC 61725203, 61732008. Yang Wang was supported by National Natural Science Foundation of China, under Grant No 61806035.

\ifCLASSOPTIONcaptionsoff
  \newpage
\fi



\bibliographystyle{IEEEtran}
\bibliography{allbib}

\begin{thebibliography}{10}
\providecommand{\url}[1]{#1}
\csname url@samestyle\endcsname
\providecommand{\newblock}{\relax}
\providecommand{\bibinfo}[2]{#2}
\providecommand{\BIBentrySTDinterwordspacing}{\spaceskip=0pt\relax}
\providecommand{\BIBentryALTinterwordstretchfactor}{4}
\providecommand{\BIBentryALTinterwordspacing}{\spaceskip=\fontdimen2\font plus
\BIBentryALTinterwordstretchfactor\fontdimen3\font minus
  \fontdimen4\font\relax}
\providecommand{\BIBforeignlanguage}[2]{{%
\expandafter\ifx\csname l@#1\endcsname\relax
\typeout{** WARNING: IEEEtran.bst: No hyphenation pattern has been}%
\typeout{** loaded for the language `#1'. Using the pattern for}%
\typeout{** the default language instead.}%
\else
\language=\csname l@#1\endcsname
\fi
#2}}
\providecommand{\BIBdecl}{\relax}
\BIBdecl

\bibitem{DG-Dropout}
T.~Xiao, H.~Li, W.~Ouyang, and X.~Wang, ``Learning deep feature representation
  with domain guided dropout for person re-identification,'' in \emph{CVPR},
  2016.

\bibitem{Deep-Embed}
L.~Wu, Y.~Wang, J.~Gao, and X.~Li, ``Deep adaptive feature embedding with local
  sample distributions for person re-identification,'' \emph{Pattern
  Recognition}, vol.~73, pp. 275--288, 2018.

\bibitem{What-and-where}
L.~Wu, Y.~Wang, X.~Li, and J.~Gao, ``What-and-where to match: Deep spatially
  multiplicative integration networks for person re-identification,''
  \emph{Pattern Recognition}, vol.~76, pp. 727--738, 2018.

\bibitem{SpindleNet}
H.~Zhao, M.~Tian, S.~Sun, J.~Shao, J.~Yan, S.~Yi, X.~Wang, and X.~Tang,
  ``Spindle net: Person re-identification with human body region guided feature
  decomposition and fusion,'' in \emph{CVPR}, 2017.

\bibitem{Part-Aligned}
L.~Zhao, X.~Li, Y.~Zhuang, and J.~Wang, ``Deeply-learned part-aligned
  representations for person re-identification,'' in \emph{ICCV}, 2017.

\bibitem{MSCAN}
D.~Li, X.~Chen, Z.~Zhang, and K.~Huang, ``Learning deep context-aware features
  over body and latent parts for person re-identification,'' in \emph{CVPR},
  2017.

\bibitem{PIE-reid}
L.~Zheng, Y.~Huang, H.~Lu, and Y.~Yang, ``Pose invariant embedding for deep
  person re-identification,'' in \emph{arXiv:1701.07732}, 2017.

\bibitem{SI-CI}
F.~Wang, W.~Zuo, L.~Lin, D.~Zhang, and L.~Zhang, ``Joint learning of
  single-image and cross-image representations for person re-identification,''
  in \emph{CVPR}, 2016, pp. 1288--1296.

\bibitem{LDAFisherVector}
L.~Wu, C.~Shen, and A.~van~den Hengel, ``Deep linear discriminant analysis on
  fisher networks: A hybrid architecture for person re-identification,''
  \emph{Pattern Recognition}, vol.~65, p. 238–250, 2017.

\bibitem{One-shot-RE-ID}
S.~Bak and P.~Carr, ``One-shot metric learning for person re-identification,''
  in \emph{CVPR}, 2017.

\bibitem{GenerativeSaliency}
H.~Wang, S.~Gong, and T.~Xiang, ``Unsupervised learning of generative topic
  saliency for person re-identification,'' in \emph{BMVC}, 2014.

\bibitem{eSDC}
R.~Zhao, W.~Ouyang, and X.~Wang, ``Unsupervised salience learning for person
  re-identification,'' in \emph{CVPR}, 2013.

\bibitem{t-LRDC}
W.-S. Zheng, S.~Gong, and T.~Xiang, ``Towards open-world person
  re-identification by one-shot group-based verification,'' \emph{TPAMI},
  vol.~38, no.~3, pp. 591--606, March 2016.

\bibitem{LSRO}
Z.~Zheng, L.~Zheng, and Y.~Yang, ``Unlabeled samples generated by gan improve
  the person re-identification baseline in vitro,'' in \emph{ICCV}, 2017.

\bibitem{CAMEL}
H.-X. Yu, A.~Wu, and W.-S. Zheng, ``Cross-view asymmetric metric learning for
  unsupervised person re-identification,'' in \emph{ICCV}, 2017.

\bibitem{OL-MANS}
J.~Zhou, P.~Yu, W.~Tang, and Y.~Wu, ``Efficient online local metric adaptation
  via negative samples for person re-identification,'' in \emph{ICCV}, 2017.

\bibitem{L1-graph}
E.~Kodirov, T.~Xiang, Z.~Fu, and S.~Gong, ``Person re-identification by
  unsupervised l1 graph learning,'' in \emph{ECCV}, 2016.

\bibitem{WangTIP15}
Y.~Wang, X.~Lin, L.~Wu, W.~Zhang, Q.~Zhang, and X.~Huang, ``Robust subspace
  clustering for multi-view data by exploiting correlation consensus,''
  \emph{IEEE Transactions on Image Processing}, vol.~24, no.~11, pp.
  3939--3949, 2015.

\bibitem{FPNN}
W.~Li, R.~Zhao, X.~Tang, and X.~Wang, ``Deepreid: Deep filter pairing neural
  network for person re-identification,'' in \emph{CVPR}, 2014.

\bibitem{Wu-TMM}
L.~Wu, Y.~Wang, J.~Gao, and X.~Li, ``Where-and-when to look: Deep siamese
  attention networks for video-based person re-identification,'' \emph{IEEE
  Transactions on Multimedia}, 2018.

\bibitem{Multi-loss}
W.~Li, X.~Zhu, and S.~Gong, ``Person re-identification by deep joint learning
  of multi-loss classification,'' in \emph{IJCAI}, 2017.

\bibitem{SPGAN}
W.~Deng, L.~Zheng, G.~Kang, Y.~Yang, Q.~Ye, and J.~Jiao, ``Image-image domain
  adaptation with preserved self-similarity and domain-dissimilarity for person
  re-identification,'' in \emph{CVPR}, 2018.

\bibitem{Re-id-ICDSC}
N.~Martinel, M.~Dunnhofer, G.~L. Foresti, and C.~Micheloni, ``Person
  re-identification via unsupervised transfer of learned visual
  representations,'' in \emph{International Conference on Distributed Smart
  Cameras}, 2017.

\bibitem{Transfer-re-id}
M.~Geng, Y.~Wang, T.~Xiang, and Y.~Tian, ``Deep transfer learning for person
  re-identification,'' in \emph{arXiv:1611.05244}, 2016.

\bibitem{GAN}
I.~J. Goodfellow, J.~Pouget-Abadie, M.~Mirza, B.~Xu, D.~Warde-Farley, S.~Ozair,
  A.~Courville, and Y.~Bengio, ``Generative adversarial nets,'' in \emph{NIPS},
  2014.

\bibitem{MTMCT}
E.~Ristani and C.~Tomasi, ``Features for multi-target multi-camera tracking and
  person re-identification,'' in \emph{CVPR}, 2018.

\bibitem{VideoPerson}
K.~Liu, B.~Ma, W.~Zhang, and R.~Huang, ``A spatio-temporal appearance
  representation for video-based person re-identification,'' in \emph{ICCV},
  2015.

\bibitem{3D-PersonVLAD}
L.~Wu, Y.~Wang, L.~Shao, and M.~Wang, ``3d personvlad: Learning deep global
  representations for video-based person re-identification,'' \emph{IEEE
  Transactions on Neural Networks and Learning Systems}, DOI:
  10.1109/TNNLS.2019.2891244, 2019.

\bibitem{NullSpace-Reid}
L.~Zhang, T.~Xiang, and S.~Gong, ``Learning a discriminative null space for
  person re-identification,'' in \emph{CVPR}, 2016.

\bibitem{UMDL}
P.~Peng, T.~Xiang, Y.~Wang, M.~Pontil, S.~Gong, T.~Huang, and Y.~Tian,
  ``Unsupervised cross-dataset transfer learning for person
  re-identification,'' in \emph{CVPR}, 2016.

\bibitem{PUL}
H.~Fan, L.~Zheng, and Y.~Yang, ``Unsupervised person re-identification:
  Clustering and fine-tuning,'' in \emph{Arxiv}, 2017.

\bibitem{DCGAN}
A.~Radford, L.~Metz, and S.~Chintala, ``Unsupervised representation learning
  with deep convolutional generative adversarial networks,'' in \emph{ICLR},
  2016.

\bibitem{Generate-chairs}
A.~Dosovitskiy, J.~T. Springenberg, M.~Tatarchenko, and T.~Brox, ``Learning to
  generate chairs, tables and cars with convolutional networks,'' in
  \emph{arXiv:1411.5928}, 2017.

\bibitem{Cycle-GAN}
J.-Y. Zhu, T.~Park, P.~Isola, and A.~A. Efros, ``Unpaired image-to-image
  translation using cycle-consistent adversarial networks,'' in \emph{ICCV},
  2017.

\bibitem{Few-shot-DA}
S.~Motiian, Q.~Jones, S.~M. Iranmanesh, and G.~Derotto, ``Few-shot adversarial
  domain adaptation,'' in \emph{NIPS}, 2017.

\bibitem{CYC-DGH}
L.~Wu, Y.~Wang, and L.~Shao, ``Cycle-consistent deep generative hashing for
  cross-modal retrieval,'' \emph{IEEE Transactions on Image Processing},
  vol.~28, no.~4, pp. 1602--1612, 2019.

\bibitem{Decaf}
J.~Donahue, Y.~Jia, O.~Vinyals, J.~Hoffman, N.~Zhang, E.~Tzeng, and T.~Darrell,
  ``Decaf: a deep convolutional activation feature for generic visual
  recognition,'' in \emph{ICML}, 2014.

\bibitem{Cross-GANs-ReID}
C.~Zhang, L.~Wu, and Y.~Wang, ``Crossing generative adversarial networks for
  cross-view person re-identification,'' \emph{Neurocomputing}, vol. 340,
  no.~7, pp. 259--269, 2019.

\bibitem{Deep-CORAL}
B.~Sun and K.~Saenko, ``Deep coral: correlation alignment for deep domain
  alignment,'' in \emph{ECCV Workshop}, 2016.

\bibitem{CORAL}
B.~Sun, J.~Feng, and K.~Saenko, ``Return of frustratingly easy domain
  adaptation,'' in \emph{AAAI}, 2016.

\bibitem{MECA}
P.~Morerio, J.~Cavazza, and V.~Murino, ``Minimal-entropy correlation alignment
  for unsupervised deep domain adaptation,'' in \emph{ICLR}, 2018.

\bibitem{Domain-confusion}
E.~Tzeng, J.~Hoffman, K.~Saenko, and T.~Darrell, ``Deep domain confusion:
  maximizing for domain invariance,'' in \emph{arXiv:1412.3474}, 2014.

\bibitem{DAN}
M.~Long and J.~Wang, ``Learning transferable features with deep adaptation
  networks,'' in \emph{ICML}, 2015.

\bibitem{Link-ICDE-2013}
G.-J. Qi, C.~C. Aggarwal, and T.~Huang, ``Link prediction across networks by
  biased cross-network sampling,'' in \emph{ICDE}, 2013, pp. 793--804.

\bibitem{Community-2013}
------, ``Online community detection in social sensing,'' in \emph{ACM
  international conference on Web search and data mining}, 2013, pp. 617--626.

\bibitem{Online-Multi-label}
X.-S. Hua and G.-J. Qi, ``Online multi-label active annotation: towards
  large-scale content-based video search,'' in \emph{ACM Multimedia}, 2008, pp.
  141--150.

\bibitem{Outlier-links-2012}
G.-J. Qi, C.~C. Aggarwal, and T.~Huang, ``On clustering heterogeneous social
  media objects with outlier links,'' in \emph{ACM international conference on
  Web search and data mining}, 2012, pp. 553--562.

\bibitem{Unified-SDA}
S.~Motiian, M.~Piccirilli, D.~A. Adjeroh, and G.~Doretto, ``Unified deep
  supervised domain adaptation and generalization,'' in \emph{ICCV}, 2017.

\bibitem{Factor-SLN-ICDM-2014}
S.~Chang, G.-J. Qi, C.~C. Aggarwal, J.~Zhou, M.~Wang, and T.~S. Huang,
  ``Factorized similarity learning in networks,'' in \emph{ICDM}, 2014, pp.
  60--69.

\bibitem{SQSS-CVPR-2016}
X.~Wang, T.~Zhang, G.-J. Qi, J.~Tang, and J.~Wang, ``Supervised quantization
  for similarity search,'' in \emph{CVPR}, 2016, pp. 2018--2026.

\bibitem{Typicality-MM-2007}
J.~Tang, X.-S. Hua, G.-J. Qi, and X.~Wu, ``Typicality ranking via
  semi-supervised multiple-instance learning.'' in \emph{ACM Multimedia}, 2007,
  pp. 297--300.

\bibitem{Flickr-Groups-IR}
J.~Wang, Z.~Zhao, J.~Zhou, H.~Wang, B.~Cui, and G.~Qi, ``Recommending flickr
  groups with social topic model,'' \emph{Inf. Retr.}, vol.~15, no. 3-4, pp.
  278--295, 2012.

\bibitem{Sim-deep-transfer}
E.~Tzeng, J.~Hoffman, T.~Darrell, and K.~Saenko, ``Simultaneous deep transfer
  across domains and tasks,'' in \emph{ICCV}, 2015.

\bibitem{ADDA}
E.~Tzeng, J.~Hoffman, K.~Saenko, and T.~Darrell, ``Adversarial discriminative
  domain adaptation,'' in \emph{CVPR}, 2017.

\bibitem{Adaptation-BPP}
Y.~Ganin and V.~Lempitsky, ``Unsupervised domain adaptation by
  backpropagation,'' in \emph{ICML}, 2015.

\bibitem{CoGAN}
M.-Y. Liu and O.~Tuzel, ``Coupled generative adversarial networks,'' in
  \emph{NIPS}, 2016.

\bibitem{M-Net}
K.~Chatfield, K.~Simonyan, A.~Vedaldi, and A.~Zisserman, ``Return of the devil
  in the details: Delving deep into convolutional nets,'' in \emph{BMVC}, 2014.

\bibitem{VGG}
K.~Simonyan and A.~Zisserman, ``Very deep convolutional networks for
  large-scale image recognition,'' in \emph{ICLR}, 2015.

\bibitem{Deep-recursive}
L.~Wu, Y.~Wang, X.~Li, and J.~Gao, ``Deep attention-based spatially recursive
  networks for fine-grained visual recognition,'' \emph{IEEE Transactions on
  Cybernetics}, vol.~49, no.~5, pp. 1791--1802, 2019.

\bibitem{ILSVRC}
O.~Russakovsky, J.~Deng, H.~Su, J.~Krause, S.~Satheesh, S.~Ma, Z.~Huang,
  A.~Karpathy, A.~Khosla, M.~Bernstein, A.~C. Berg, and L.~Fei-Fei, ``Imagenet
  large scale visual recognition challenge,'' in \emph{arXiv preprint,
  arXiv:1409.0575}, 2014.

\bibitem{Conditional-GAN}
P.~Isola, J.~Zhu, T.~Zhou, and A.~Efros, ``Image-to-image translation with
  conditional adversarial networks,'' in \emph{arXiv:1611.07004}, 2016.

\bibitem{Gray2007Evaluating}
D.~Gray, S.~Brennan, and H.~Tao, ``Evaluating appearance models for
  recognition, reacquisition, and tracking,'' in \emph{IEEE International
  Workshop on Performance Evaluation of Tracking and Surveillance}, 2007.

\bibitem{Market1501}
L.~Zheng, L.~Shen, L.~Tian, S.~Wang, J.~Wang, and Q.~Tian, ``Scalable person
  re-identification: A benchmark,'' in \emph{ICCV}, 2015.

\bibitem{DukeMTMC}
E.~Ristani, F.~Solera, R.~S. Zou, R.~Cucchiara, and C.~Tomasi, ``Performance
  measures and a data set for multi-target, multi-camera tracking,'' in
  \emph{ECCV Workshop on Benchmarking Multi-Target Tracking}, 2016.

\bibitem{MarketDetector}
B.~Huang, J.~Chen, Y.~Wang, C.~Liang, Z.~Wang, and K.~Sun, ``Sparsity-based
  occlusion handling method for person re-identification,'' in \emph{Multimedia
  Modeling}, 2015.

\bibitem{DCSL}
Y.~Zhang, X.~Li, L.~Zhao, and Z.~Zhang, ``Semantics-aware deep correspondence
  structure learning for robust person re-identification,'' in \emph{IJCAI},
  2016.

\bibitem{SPReID}
M.~M. Kalayeh, E.~Basaran, M.~Gokmen, M.~E. Kamasak, and M.~Shah, ``Human
  semantic parsing for person re-identification,'' in \emph{CVPR}, 2018.

\bibitem{SVDNet}
Y.~Sun, L.~Zheng, W.~Deng, and S.~Wang, ``Svdnet for pedestrian retrieval,'' in
  \emph{ICCV}, 2017.

\bibitem{DPFL}
Y.~Chen, X.~Zhu, and S.~Gong, ``Person re-identification by deep learning
  multi-scale representations,'' in \emph{ICCV workshops}, 2017.

\bibitem{LOMOMetric}
S.~Liao, Y.~Hu, X.~Zhu, and S.~Z. Li, ``Person re-identification by local
  maximal occurrence representation and metric learning,'' in \emph{CVPR},
  2015, pp. 2197--2206.

\bibitem{ResNet}
K.~He, X.~Zhang, S.~Ren, and J.~Sun, ``Deep residual learning for image
  recognition,'' in \emph{CVPR}, 2016.

\bibitem{Match-nets}
O.~Vinyals, C.~Blundell, T.~Lillicrap, K.~Kavukcuoglu, and D.~Wierstra,
  ``Matching networks for one shot learning,'' in \emph{arXiv:1606.04080},
  2016.

\bibitem{Model-regress}
Y.~Wang and M.~Hebert, ``Learning to learn: model regression networks for easy
  small sample learning,'' in \emph{ECCV}, 2016.

\end{thebibliography}

\end{document}